\documentclass[journal]{IEEEtai}
\let\labelindent\relax
\usepackage{array}
\usepackage{textcomp}
\usepackage{stfloats}
\usepackage{url}
\usepackage{verbatim}
\usepackage{booktabs} 
\usepackage{graphicx}
\usepackage{epstopdf}
\usepackage{subfigure}
\usepackage{amsmath,amssymb,amsfonts}
\usepackage{xcolor}
\usepackage{enumerate}
\usepackage{amsthm}
\usepackage{diagbox}
\usepackage{enumitem} 
\usepackage{color,colordvi}
\usepackage{epsfig}
\usepackage{mdwlist}
\usepackage{makecell}
\usepackage{multirow,multicol}
\usepackage{setspace}
\usepackage{float}
\usepackage{fancyhdr}
\usepackage[normalem]{ulem}
\usepackage{makecell}
\usepackage{fancyhdr}
\usepackage{flushend}
\usepackage{algorithm}
\usepackage{algorithmic}
\usepackage{cite}
\usepackage{multirow}
\usepackage{soul}
\usepackage[colorlinks,linkcolor=green,citecolor=green]{hyperref}
{}
{}
{}

\begin{document}
\title{Towards Generalized Multi-stage Clustering: Multi-view Self-distillation}
\author{{Jiatai Wang \href{https://orcid.org/0000-0001-7373-7706}{\includegraphics[scale=0.08]{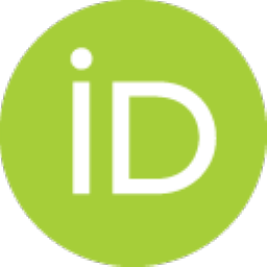}}}, 
        Zhiwei~Xu,
        Xin~Wang,
        Tao Li,~\IEEEmembership{Member,~IEEE}
\thanks{This work  was supported by the National Science Foundation of China (61962045, 62062055, 61902382, 61972381),  the Science and Technology Planning Project of Inner Mongolia Autonomous Region (2019GG372).
(Corresponding Author: Zhiwei Xu.)
}
\IEEEcompsocitemizethanks{

\IEEEcompsocthanksitem Jiatai Wang and Tao Li are with the College of Computer Science, Nankai University, Tianjin 300350, China, and also with the Xingchuang Haihe Laboratory, Tianjin 300459, China (e-mail: wangjiatai@hl-it.cn; litao@nankai.edu.cn).
\IEEEcompsocthanksitem Zhiwei Xu is with Xingchuang Haihe Laboratory, Tianjin, China, 300459, while visiting at Institute of Computing Technology, Chinese Academy of Sciences, Beijing, China, 100190 (E-mail: xuzhwei2001@ict.ac.cn).\protect
\IEEEcompsocthanksitem Xin Wang is with the Department of Electrical and Computer Engineering, Stony Brook University, New York, U.S.A. 11794 (E-mail: x.wang@stonybrook.edu).
}}

\markboth{IEEE TRANSACTIONS ON NEURAL NETWORKS AND LEARNING SYSTEMS, VOL, AUGUST, 2022}%
{Multi-view Clustering with Self-distillation~}
\maketitle

\begin{abstract}
Existing multi-stage clustering methods independently learn the salient features from multiple views and then perform the clustering task. Particularly, multi-view clustering (MVC) has attracted a lot of attention in multi-view or multi-modal scenarios. MVC aims at exploring common semantics and pseudo-labels from multiple views and clustering in a self-supervised manner. However, limited by noisy data and inadequate feature learning, such a clustering paradigm generates overconfident pseudo-labels that mis-guide the model to produce inaccurate predictions. Therefore, it is desirable to have a method that can correct this pseudo-label mistraction in multi-stage clustering to avoid the bias accumulation.  To alleviate the effect of overconfident pseudo-labels and improve the generalization ability of the model, this paper proposes a novel multi-stage deep MVC framework where multi-view self-distillation (DistilMVC) is introduced to distill dark knowledge of label distribution. Specifically, in the feature subspace at different hierarchies, we explore the common semantics of multiple views through contrastive learning and obtain pseudo-labels by maximizing the mutual information between views. Additionally, a teacher network is responsible for distilling pseudo-labels into dark knowledge, supervising the student network and improving its predictive capabilities to enhance the robustness. Extensive experiments on real-world multi-view datasets show that our method has better clustering performance than state-of-the-art methods.

\end{abstract}


\section{INTRODUCTION}\label{sec1}

\IEEEPARstart{T}{\lowercase{raditional}} clustering methods \cite{13,14,2,16,17,18,19,20,IMVTSC-MVI} have been used with specific machine learning techniques in various tasks. Among them, clustering algorithms \cite{6,32,47} based on deep learning have emerged due to their powerful generalization capability and scalability. These algorithms jointly learn the parameters of some specific neural networks and assign the features extracted to clusters. Among them, one-stage deep clustering methods \cite{21,eamc,SiMVC} work end-to-end for feature learning, and are easy to lock in low-level features. 
On the other hand, the multi-stage deep clustering method \cite{MVC-LFA,MFLVC} performs multiple rounds of feature extraction under the supervision of the pseudo-labels obtained through self-learning, where the labels are used to guide the training of a prediction model for clustering. The overall process of multi-stage deep clustering fits exactly into the self-supervised paradigm of model training guided by the intrinsic structure of data, which helps to achieve enhanced feature learning and clustering performance. According to Cover's theorem \cite{cover1965}, complex data are more likely to be linearly separable when they are  projected to a high dimensional representation space, and this theory provides a base for the feasibility of such pseudo-label-based training. 
The pseudo-labels learnt are used as a priori or self-supervised signal to guide training of clustering model \cite{55,MFLVC,22,32,55,56}. Recently, multi-stage clustering methods have become a focus of research \cite{32}.
\begin{figure}[t!]
\centering
\includegraphics[width=0.88\linewidth]{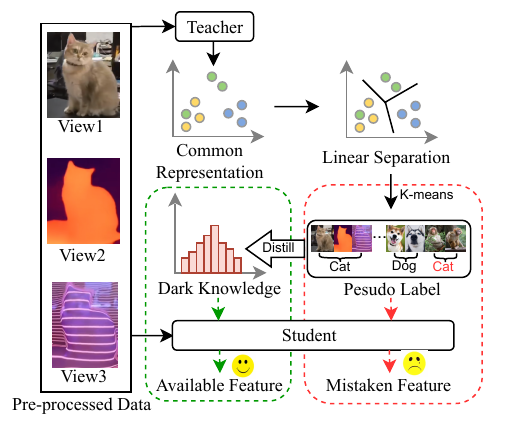}
\caption{Overconfident pseudo-labels used in MVC and their distillation. 
The pre-processed multi-view data instances are learned to achieve common representation of views. However, pseudo-labels obtained from common representation learning are often overconfident for this multi-view scenario. Distillation after labelling, obtains dark knowledge, a new self-supervised signal that contains richer semantic information compared to pseudo-labels, can better guide the multi-stage clustering and significantly improve the quality of clustering.}
\label{fig0}
\end{figure}

Data in real world are mostly collected from different (types of) sensors or feature extractors. Multi-View Clustering (MVC), one of the multi-stage clustering problems, has been proposed to explore the common semantics among different views and investigate the effectiveness of pseudo-labeling for self-supervision \cite{7,8,MNIST-USPScomic,12}. 
However, MVC suffers from some drawbacks and constraints when applied to multi-modal or multi-views.
Although samples of different views include more features, the distance measures in the high-dimensional representation space of multi-views are no longer reliable due to dimensional catastrophes, imbalanced data distribution, and noise pollution~\cite{motivation1, motivation2, motivation3}, leading to the overconfidence in K-means or other basic clustering methods and thus biased pseudo-labelling.
If a pseudo-label is obtained directly with K-means, the intra-cluster and inter-cluster associations are ignored, leading to the overconfidence in the pseudo-label (i.e., low entropy prediction) \cite{58}.
Thus, it is a challenge to avoid the damaging impact of false pseudo-labels during feature learning and correct the inaccurate bootstrapping \cite{SiMVC,MFLVC}.

To address this challenge, we study multi-stage deep MVC methods comprehensively and find that the use of knowledge distillation can considerably enhance model performance in both supervised and unsupervised settings~\cite{FT,DeiT,KD-RP,63,64}. In this case, a teacher network transfers implicit information (dark knowledge) \cite{38} to the student network so that it can distinguish similarities and differences among samples. More specifically, for unsupervised MVC tasks, the success of self-distillation even with a weak teacher is not solely due to the knowledge shared by the teacher, but rather due to the regularization of the distilled knowledge~\cite{76,77}.
Based on these observations, we propose a novel multi-stage deep MVC framework based on multi-view self-distillation (DistilMVC), which can distill pseudo labels into the dark knowledge which serves as a new self-supervised signal to guide the feature learning (see Fig. \ref{fig0}). DistilMVC projects multi-view instances into hierarchical feature spaces and ensures the consistency of multi-view representation learning. More specifically, we introduce KL divergence and self-distillation structures to replace the overconfident pseudo-labeling with dark knowledge of multiple hierarchies, and introduce a contrastive loss to learn features by maximizing the mutual information of different views in different hierarchies of the latent space. Our contributions can be summarized as:
\begin{itemize}
    \item We explore the use of knowledge distillation in MVC, and propose a multi-view self-distillation technology that transforms overconfident pseudo-labels into dark knowledge, reducing the impact of false pseudo-labels on multi-view feature learning. As dark knowledge contains essential hierarchical information that is not included in pseudo-labels, using it as a supervision indicator can generalize the multi-view representation learning.

    \item We propose a contrastive method to learn multi-view semantics in feature spaces from different hierarchies. In a low-dimensional latent space, we directly maximize the mutual information with invariant information clustering, and in a high-dimensional subspace, we raise the lower bound of mutual information according to the fixed point related to the scale of negative samples. This can accordingly improve the self-supervised learning multi-view representation performance for MVC. 
    
    \item Based on the proposed multi-view self-distillation technology, we introduce a new multi-stage framework, which uses the dark knowledge instead of pseudo-labels as a supervision indicator and thus generalize MVC capability.
    
    \item Experiments on eight real-world image datasets demonstrate that DistilMVC outperforms state-of-the-art clustering performance and can achieve strong robustness. 
\end{itemize}

To our best knowledge, DistilMVC is the first method to incorporate knowledge distillation into self-supervised feature learning of MVC, providing a novel solution for high-quality multi-view clustering method.
This allows MVC models to be embedded into the physical world to learn more consistent representation in broad scenarios in a self-supervised way.

\section{RELATED  WORK}\label{sec2}
In this section, we briefly review three lines of related work, deep multi-view clustering, contrastive learning, and knowledge distillation.

\subsection {Deep Multi-View Clustering}
As the mainstream type of enhanced multi-stage clustering approaches, multi-view clustering (MVC) has attracted increasingly wide attention from researchers.
Traditional MVC methods \cite{13,14,2,16,17,18,19,20,IMVTSC-MVI,69,70} have a number of limitations, including high complexity, slow speed, and difficult deployment in real-world scenarios. In recent years, deep learning-based multi-view clustering methods \cite{21,22,24,25,26,27,50,29,MFLVC,71,72,73} have received more and more attention. They exploit the excellent representation ability from multi-view data latent clustering patterns. Such methods can be roughly divided into two categories, namely one-stage and multi-stage methods. Most of the one-stage methods \cite{21,eamc,SiMVC} are designed to work end-to-end. Synchronizing feature learning and clustering taken by this kind of methods can effectively reduce the multi-stage error accumulation, and better support  streaming data processing. The multi-stage methods \cite{22,MVC-LFA,MFLVC} follow the self-supervised learning paradigm, first pre-training for feature learning and then fine-tuning according to different proxy tasks or algorithms. One-stage methods are likely to latch onto low-level  features because of their dependence on initialization, so the multi-stage method with pretraining usually has better performance in providing higher accuracy.

The proposed DistilMVC is a multi-stage MVC framework that requires pretraining to obtain rich prior knowledge, which avoids relying on low-level features in the clustering learning process. Almost all MVC methods do not take into account the inaccurate guidance from the use of pseudo-labels and thus suffer from model degradation. To address this issue, we replace pseudo-labels with dark knowledge from the perspective of knowledge distillation.

\subsection {Contrastive Learning}
Contrastive learning \cite{30,33,35,36,37} is an essential method for unsupervised learning \cite{31}. Its major goal is to maximize feature space similarity between positive samples while reducing the distance between negative samples. In the field of computer vision, contrastive learning methods have produced excellent results~\cite{32}. For example, SimClR\cite{30} or MoCo\cite{33}  minimize the InfoNCE loss function \cite{34} to maximize the lower bound of mutual information. Since the processing of negative samples is very cumbersome, the follow-up work, BYOL \cite{35}, SimSiam \cite{36}, and DINO \cite{37} have successfully transformed the contrastive task into a prediction task without defining negative samples and achieved amazing results.

Previous work simply constructs positive and negative samples based on data augmentation. Although these studies have shown that consistency could be learned by maximizing the mutual information of different views, they ignore the mutual information at different hierarchies. In contrast, our method aims to learn shared semantics from  multiple views. DistilMVC first constructs two independent subspaces and defines positive and negative samples according to the feature matrix in each subspace respectively, and then uses the InfoNCE  loss to maximizes the lower bound of mutual information of different views. 

\subsection {Knowledge Distillation}
Knowledge Distillation (KD) is a model compression method in which a smaller student model relies on a pretrained teacher model to obtain performance close to or even surpassing the teacher model. In order to help students learn more semantic information, minimizing the loss of the output class probability (soft label) of the teacher model \cite{38} can make the soft label contain rich dark knowledge. 

The differences between this work and existing knowledge distillation studies are as below. DistilMVC adopts a self-distillation \cite{40,41,42,43} method that does not require a pretrained model of the teacher network, nor does it need to detach the gradient of the teacher network. In DistilMVC, the student network and the teacher network do collaborative training, and the teacher network relies on the momentum update \cite{33} of the student network parameters, which is conducive to maintaining consistent semantic information for high-dimensional features. The proposed method extracts the dark knowledge from high-dimensional features, supervises the learning of the student network, and improves the generalization ability of the model \cite{52}. To the best of our knowledge, this is the first work that applies knowledge distillation to multi-view clustering, which optimizes pseudo-labels quality and improves the clustering performance.

\section{Revisiting Knowledge Distillation Used in Multi-stage Learning Tasks}

A multi-stage deep learning task \cite{32, RetinaNet}, including Multi-stage MVC \cite{MVC-LFA,MFLVC,57}, leverages K-means and other basic clustering methods\cite{63} to converts high-dimensional features into pseudo-labels to guide learning tasks.
However, the distance measures in high-dimensional spaces are not reliable due to dimensional catastrophes, imbalanced data distribution, and noise pollution~\cite{motivation1, motivation2, motivation3}, leading to the overconfidence in K-means or other basic clustering methods and thus the biased pseudo-labelling. As the noise accumulates, the obtained pseudo-labels \cite{55,58} lose intra-cluster and inter-cluster associations, degrading the model prediction performance (low-entropy prediction).


\begin{figure}[h]
\centering
\includegraphics[width=1.164\linewidth]{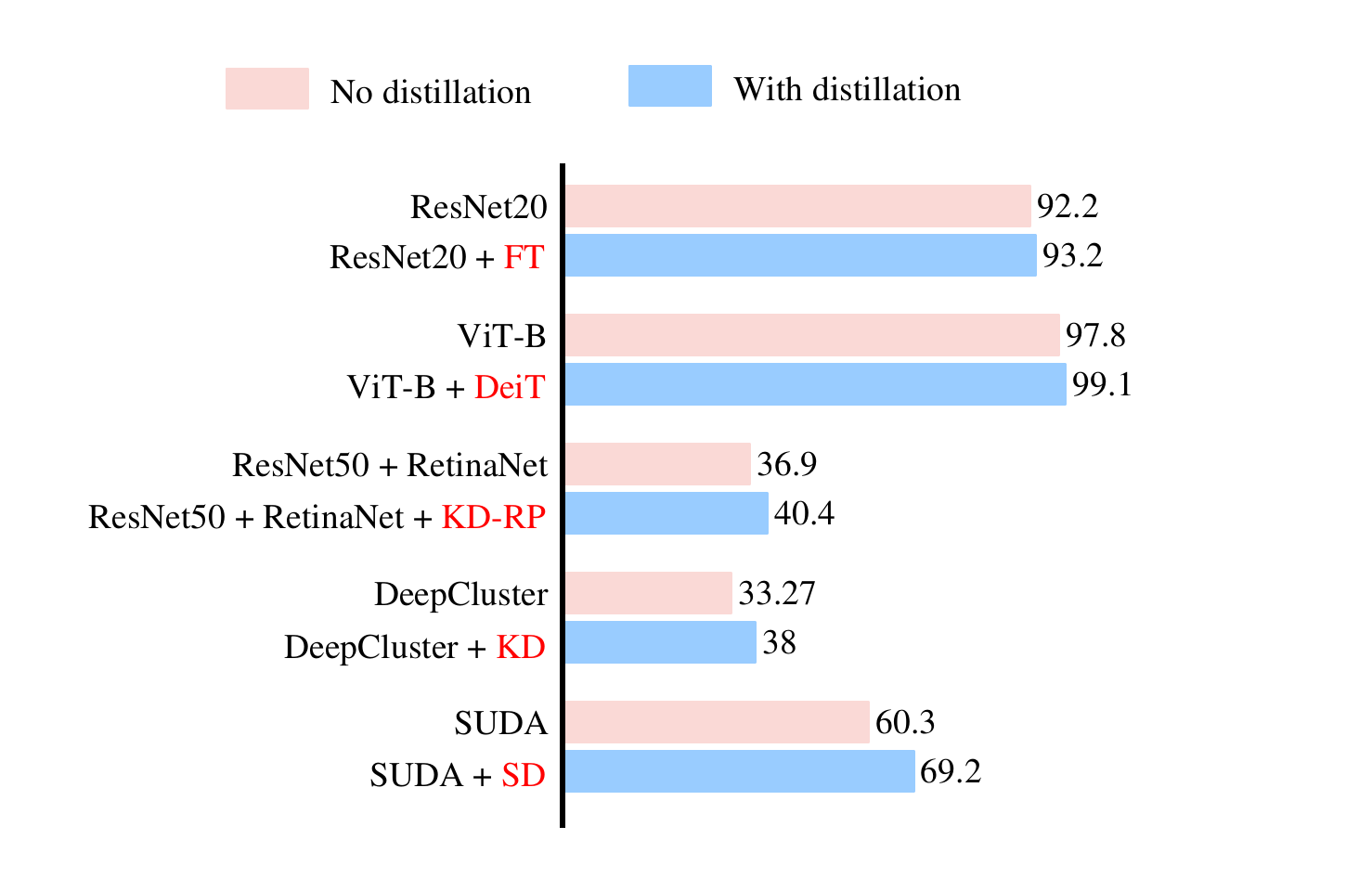}
\caption{Comparison of learning performance of visual tasks with or without distillation. In this figure, we display the performance improvements of different feature extractors with an additional distillation processes. The performance improves in the cases of using the convolution-based ResNet \cite{ResNet}, the self-attention-based ViT \cite{ViT}, the object detection network RetinaNet \cite{RetinaNet}, 
the CNN based Deep Clustering\cite{74}, and the unsupervised domain adaptation\cite{75}.}
\label{fig:distillation}
\end{figure}

Inspired by the fact that knowledge distillation is feasible to tackle low-entropy prediction problems\cite{58,60}, we explore the use of knowledge distillation in multi-stage learning tasks. 
More specifically, we perform five experiments, three of which are supervised tasks and two are unsupervised tasks, and incorporate a knowledge distillation method into each task.
The specific experimental settings are shown in Table \ref{table00}.
The corresponding distillation methods are as follows: 
1) FT \cite{FT} uses convolutional operations to transfer dark knowledge; 
2) DeiT \cite{DeiT} proposes the distillation token and uses its representation with the teacher model's dark knowledge to compute the distillation loss;
3) KD-RP \cite{KD-RP} exploits the differences in student and teacher networks to guide dark knowledge distillation; 4) KD \cite{63}  provides additional information about semantic similarity to model learning through the use of dark knowledge generated by self-distillation; 5) SD \cite{64} exploits self-distillation to learn effective representations to group point clouds in the target domain.

\begin{table*}[h]
\caption{Backbone settings for different vision tasks and their corresponding improved knowledge distillation methods.}
\label{table00}
\renewcommand\arraystretch{1}
\centering
\setlength{\tabcolsep}{1.4mm}{
\begin{tabular}{lllll}
\hline
 & Dataset & Backbone                                              
& \begin{tabular}[c]{@{}l@{}}Distillation\end{tabular} & Metrics\\ \hline
\begin{tabular}[c]{@{}l@{}}Image classification\end{tabular}
&\begin{tabular}[c]{@{}l@{}}CIFAR10\cite{CIFAR10}\end{tabular}
&\begin{tabular}[c]{@{}l@{}}ResNet20\cite{ResNet} \end{tabular}      & FT
& Accuracy                                                           \\
\begin{tabular}[c]{@{}l@{}}Image classification\end{tabular} & \begin{tabular}[c]{@{}l@{}}CIFAR10\cite{CIFAR10}\end{tabular} & \begin{tabular}[c]{@{}l@{}}ViT-B\cite{ViT} \end{tabular}             & DeiT-B
& Accuracy                                                           \\
\begin{tabular}[c]{@{}l@{}}Object detection\end{tabular} & \begin{tabular}[c]{@{}l@{}}CoCo\cite{CoCo} \end{tabular} & \begin{tabular}[c]{@{}l@{}}
ResNet20+RetinaNet\cite{ResNet,RetinaNet}\end{tabular} 
& KD-RP
& \begin{tabular}[c]{@{}l@{}} Average Precision\end{tabular} 
\\
\begin{tabular}[c]{@{}l@{}} Image classification \end{tabular} & \begin{tabular}[c]{@{}l@{}} CIFAR10\cite{CIFAR10}\end{tabular} & \begin{tabular}[c]{@{}l@{}} Deep Cluster\cite{65} \end{tabular} 
& KD
& Accuracy                                                     \\
\begin{tabular}[c]{@{}l@{}} 3D vision classification \end{tabular} 
&\begin{tabular}[c]{@{}l@{}} PointDA-10 \cite{66}\end{tabular} &\begin{tabular}[c]{@{}l@{}} PointNet+DGCNN\cite{67,68} \end{tabular}
& SD
& Average Precision                                            \\ \hline
\end{tabular}
}
\end{table*}

\begin{figure*}[htbp]
  \begin{center}
  \includegraphics[width=0.82\textwidth]{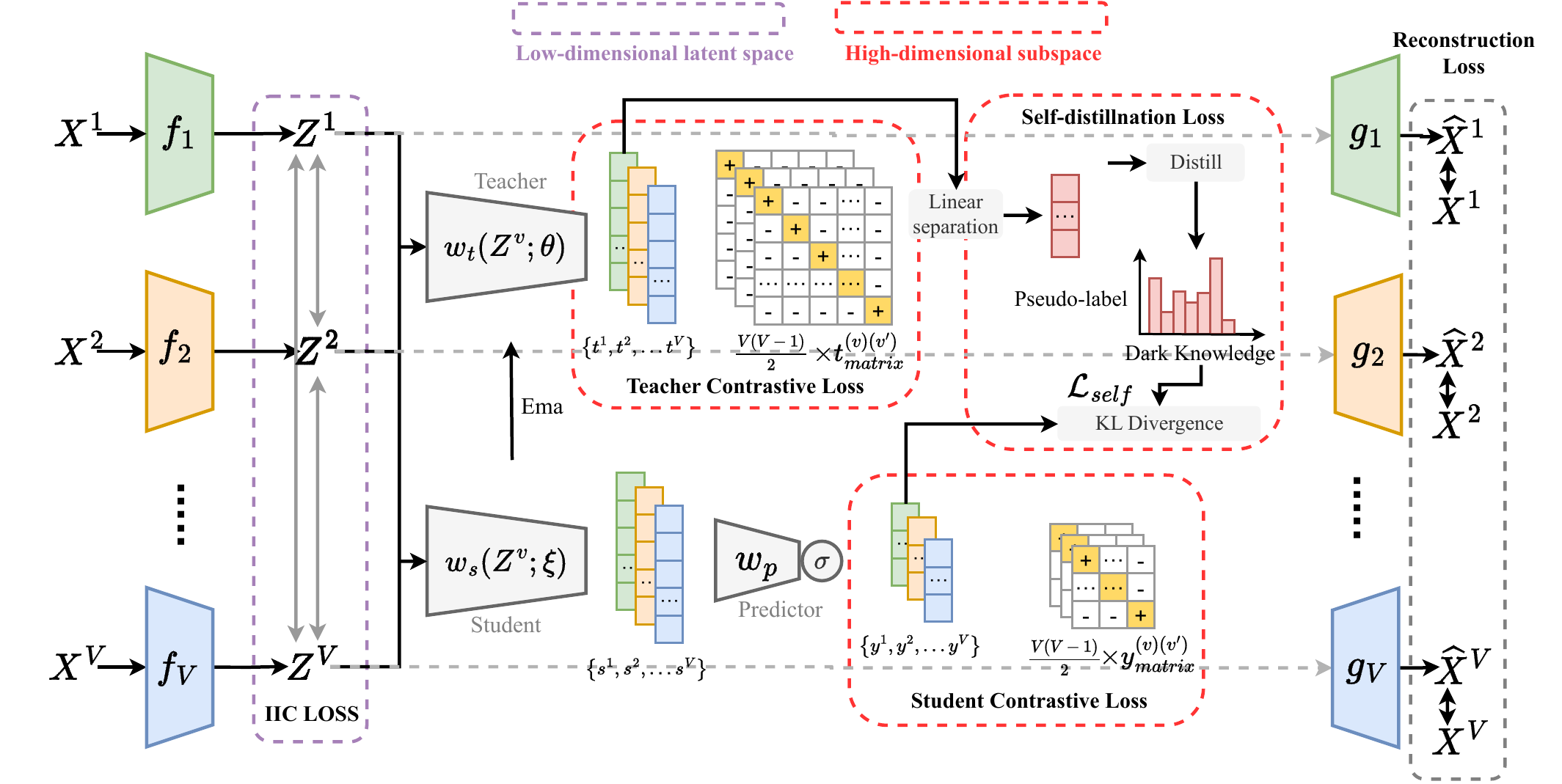}
  \caption{The framework of the proposed DistilMVC. The encoder $f_v$ and decoder $g_v$ learn latent representation $Z^{v}$ for the $v$-th view by reconstructing $X^{v}$ (Section \ref{3.3}). The student network $w_{s}$ and the teacher network $w_{t}$ capture hierarchical representations through contrastive learning in their subspaces, and the latent representations $\left\{Z^{1}, Z^{2}, ...Z^{v}\right\}$ maximize the mutual information pairwise (Section \ref{3.4}). The probability distribution of the obtained features of the student network will be compared with the dark knowledge of the teacher network to calculate KL divergence (Section \ref{3.5}), where "Ema" denotes exponential moving average, and the teacher network is updated with momentum by the parameters of the student network.
 }\label{fig1}
  \end{center}
\end{figure*}

The experimental results are shown in Fig. \ref{fig:distillation}, with the corresponding distillation methods highlighted in red.
 The five tasks can all improve the performance of their backbone networks after exploiting the Knowledge distillation.
Compared with pseudo labels, dark knowledge from the teacher contains the similarity information between classes~\cite{76}.
With the incorporation of self-distillation, a weak teacher with much lower accuracy than students can still significantly improve the clustering accuracy of students. The success of self-distillation even with a weak teacher is not solely due to the shared similarity information between classes, but rather due to the regularization of the distilled knowledge \cite{76}. 
This demonstrates that dark knowledge of knowledge distillation plays a positive role in different learning tasks.

In the next section, we consider this observation and leverage knowledge self-distillation in Multi-stage MVC.

\section{The Proposed Multi-view Clustering with Self-distillation Method}\label{sec3}
Multi-view data introduces more features, and thus over-confident pseudo-labels is poor to represent these features accompanied by more noise, which results in existing multi-stage clustering methods are difficult to adapt to this multi-view clustering scenario. 

To solve the above-mentioned issues and alleviate the overconfidence of pseudo-labels while learning the common semantics of different views, we propose a novel technique, the multi-view distillation technique. Its contrastive method to learn multi-view semantics from different hierarchies is present in the first place. Then, we incorporate this technique into a novel multi-stage MVC framework (DistilMVC).

\subsection{Framework Overview}
\label{3.2}

Given a multi-view dataset $\mathcal{X} = \left\{X^{v} \in \mathbb{R}^{N_{} \times D_{v}}\right\}_{v=1}^{V}$ , where each view takes $N$ samples. $V$ denotes the number of views, $v \in \left\{1,...V\right\}$. $D_{v}$ denotes the dimension of the $v$-th view sample $X^{v}$, and $k$ is the number of categories to cluster.  
We show the framework of our proposed DistilMVC in Fig. \ref{fig1}. To reconstruct the views and build the feature space, DistilMVC is equipped with an autoencoder for each view, and the encoder and decoder for the view $v$ are denoted by $f_{v}$ and $g_{v}$. DistilMVC is a self-supervised algorithm. Self-supervised learning is an unsupervised learning scheme because it follows the criteria that no labels are given but constructs pseudo-labels from the data itself. Combined with the idea of distillation, the whole self-supervised learning process consists of a student network and a teacher network, simultaneously learning a common representation of multiple views, with the teacher network providing optimization directions for the student network. Two student networks ($w_{s}$ and $w_{p}$) and a teacher ($w_{t}$) network are shared by all views and applied to extract multi-view features and project the original features to the feature spaces of different hierarchies.
The predictor $w_{p}$ converts the features of $w_{s}$ into probability distributions and uses them as soft labels for distillation. DistilMVC constructs two high-dimensional subspaces and a low-dimensional latent space, and learns common semantics by maximizing the mutual information of the feature spaces with different hierarchies. Specifically, the student network and the teacher network will construct two independent high-dimensional subspaces and indirectly improve the lower boundary of mutual information through contrastive learning in their respective subspaces. At the same time, we introduce invariant information clustering \cite{23} to directly maximize the mutual information of low-dimensional features. The teacher network will linearly separate the learned high-dimensional features into pseudo-labels. To combat the overconfidence of pseudo-labels, we designed a self-distillation algorithm. Specifically, the teacher network outputs $k$-dimensional features and converts one-dimensional pseudo-labels into $k$-dimensional dark knowledge by adjusting the temperature and adding a Softmax activation function. The dark knowledge obtained by the final distillation is used as the ground truth, and the KL divergence is used to measure its similarity to the output of the student network.

\subsection{Reconstruction Loss}
\label{3.3}
Deep autoencoders can capture the salient features of data and have applications in many unsupervised domains \cite{45,46}. Therefore, we minimize
\begin{equation}
\label{eq1}
\begin{aligned}
\mathcal{L}_{r e c}&=\sum_{v=1}^{V} \sum_{n=1}^{N}\left\|X_{n}^{v}-g^{v}\left(f^{v}\left(X_{n}^{v}\right)\right)\right\|_{2}^{2}\\&=\sum_{v=1}^{V} \sum_{n=1}^{N}\left\|X_{n}^{v}-g^{v}\left(Z_{n}^{v}\right)\right\|_{2}^{2}
\end{aligned}
\end{equation}
to enable the autoencoder to convert heterogeneous multi-view data into a cluster-friendly latent representation $Z^{v}$. For the $v$-th view, $X_{n}^{v}$ represents the $n$-th feature vector. The learned latent representation is defined as $Z^{v}$, and $Z_{n}^{v}$ denotes the $n$-th latent representation. $\hat{X}^{v}$ is the reconstructed view of $Z^{v}$. This design can make the autoencoder maintain the respective diversity of views, avoid the trivial solution, and prevent model collapse, which is the basis for improving the performance of multi-view clustering.

\subsection{Contrastive Loss}
\label{3.4}
For the model to perform feature learning effectively, the teacher network and the student network project the low-dimensional representation $\left\{Z^{1}, Z^{2}, ...Z^{v}\right\}$ into the higher-dimensional spaces $\left\{t^{1}, t^{2}, ...t^{v}\right\}$ and $\left\{y^{1}, y^{2}, ...y^{v}\right\}$ at different hierarchies, respectively. To enable effective feature learning at different hierarchies, we take the following procedures:
(1) Optimizing $\mathcal{L}_{stu}$ and $\mathcal{L}_{tea}$ to indirectly raise the lower bound of mutual information between views; 
(2) Optimizing $\mathcal{L}_{IIC}$ to directly maximize the mutual information between views. We propose an objective function for learning common semantics:
\begin{equation}
\label{eq2}
\begin{aligned}
\mathcal{L}_{con}=\mathcal{L}_{stu}+\mathcal{L}_{tea}+\mathcal{L}_{IIC}.
\end{aligned}
\end{equation}
Each component of this objective function will be described in details below.

\subsubsection{Student Contrastive Loss}
\begin{figure}[htbp]
\centering
\includegraphics[width=0.8\linewidth]{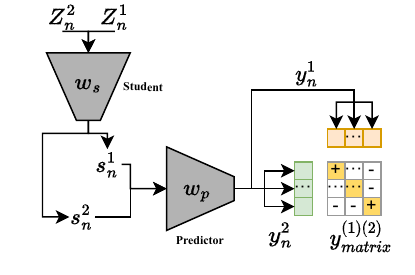}
\caption{Calculation of student contrastive loss. A group of shared deep neural networks $w_{s}$ and $w_{p}$ are used to extract features from different views. The predictor $w_{p}$ is used to project the features into high-dimensional subspaces where $y_{n}^{1}$ and $y_{n}^{2}$ denote pseudo-labels generated by Softmax operations in this contrastive learning. The feature matrix 
$y^{(1)(2)}_{matrix}$ is obtained by multiplying $y_{n}^{1}$ and $y_{n}^{2}$, to learn common semantics. }
\label{fig2}
\end{figure}

Fig. \ref{fig2} shows how contrastive learning is used in  the student network in the example case of $V=2$. Given a batch of $n$ ($Z_{n}^{1}, Z_{n}^{2}$) pairs,  a student network is trained to predict which of the $n \times n$ possible ($Z_{n}^{1}, Z_{n}^{2}$) pairings across a batch actually occurred.
To do this, $w_{p}$ learns the multi-view embedding space feature matrix $y_{matrix}^{(v)(v^{\prime})}$ by maximizing the cosine similarity of $y_{n}^{1}$ and $y_{n}^{2}$ of $n$ positive sample pairs on the diagonal while simultaneously minimizing the cosine similarity of the embeddings of $(n^{2}-n)$ negative sample pairs. The pairwise similarity in the feature matrix is measured by cosine similarity as
\begin{equation}
\label{eq3}
\begin{aligned}
cos\left(y_{n}^{v}, y_{m}^{v^{\prime}}\right)=\frac{\left(y_{n}^{v}\right)(y_{m}^{v^{\prime}})^{\top}}{\left\|y_{n}^{v}\right\|\left\|y_{m}^{v^{\prime}}\right\|},
\end{aligned}
\end{equation}
where $n, m \in[1, N]$, $v, v^{\prime} \in[1, V]$ and $v \ne v^{\prime}$. In order to optimize the pairwise similarity, without loss of generality, given the sample pairs $y_{n}^{v}$ and $y_{n}^{v^{\prime}}$, we optimize the symmetric cross entropy loss:
\begin{equation}
\label{eq4}
\begin{aligned}
&\ell_{y}^{(v)(v^{\prime})}=-\frac{1}{2 N} \sum_{n=1}^{N}\log \\&\frac{\exp \left(cos\left(y_{n}^{v}, y_{n}^{v^{\prime}}\right) / \tau_{s}\right)}{\sum_{m=1}^{N}\left[\exp \left(cos\left(y_{n}^{v}, y_{m}^{v}\right) / \tau_{s}\right)+\exp \left(cos\left(y_{n}^{v}, y_{m}^{v^{\prime}}\right) / \tau_{s}\right)\right]},
\end{aligned}
\end{equation}
where $\tau_{s}$ is the student network temperature parameter that controls the softness of the distribution. Since we wish to identify all positive pairs of the entire dataset, the contrastive loss of sample pairs $s_{n}^{v}$ and $s_{n}^{v^{\prime}}$ needs to be computed on all views, which we extend to $V \ge 2$ below:
\begin{equation}
\label{eq5}
\begin{aligned}
\mathcal{L}_{stu}=\sum_{v=1}^{V} \sum_{v \neq v^{\prime}} \ell_{y}^{(v)(v^{\prime})}-H(Y).
\end{aligned}
\end{equation}
In Equation \ref{eq5}, we add an additional entropy balance term
\begin{equation}
\label{eq6}
\begin{aligned}
H(Y)=-\sum_{v=1}^{V}\left[P\left(y_{n}^{v}\right) \log P\left(y_{n}^{v}\right)+P\left(y_{n}^{v^{\prime}}\right) \log P\left(y_{n}^{v^{\prime}}\right)\right],
\end{aligned}
\end{equation}
This regularization term avoids the trivial solution and prevents all sample points from clustering into the same class.

\begin{figure}[htbp]
\centering
\includegraphics[width=0.75\linewidth]{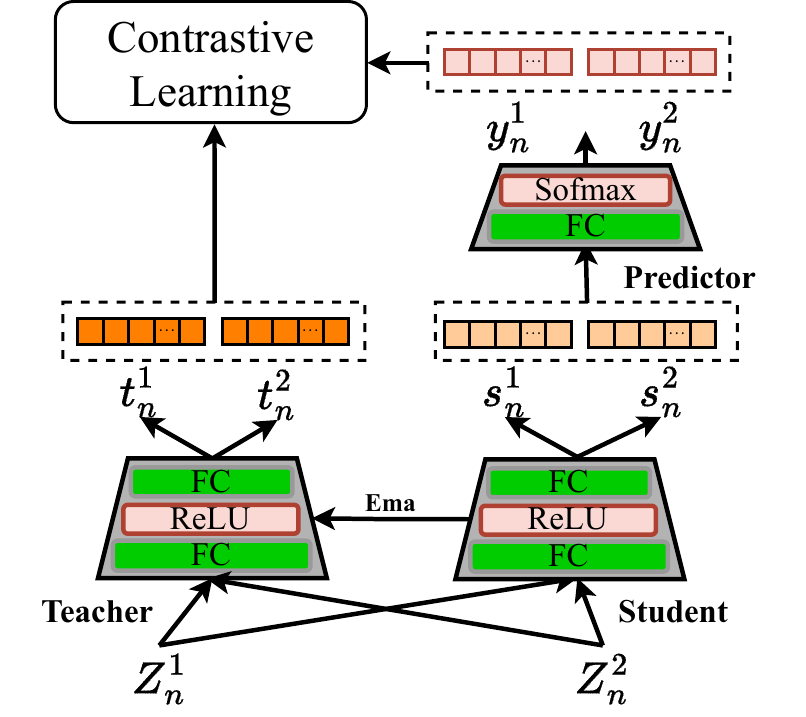}
\caption{Illustration of the model structure of the student network and teacher network.}
\label{fig3}
\end{figure}

\subsubsection{Teacher Contrastive Loss}
As seen in Fig. \ref{fig3}, both the teacher network and the student network use the same feature learning methods. The only distinction is that the teacher network doesn't require a regularization term to prevent model collapse. The goal of the teacher network is to provide a supervised signal for the optimization of the student network while providing high-dimensional features $\left\{t^{1}, t^{2}, ...t^{v}\right\}$ for linear separation. However, if the regularization term is added, it would smooth the original distribution of teacher network features, weakening the linear separability. Similar to the student network, the use of contrastive learning helps better fit the probability distribution and learn the mutual information of different hierarchies. Similarly, we give the sample pair $t_{n}^{v}$ and $t_{n}^{v^{\prime}}$ to optimize the symmetric cross-entropy loss as:
\begin{equation}
\label{eq7}
\begin{aligned}
&\ell_{t}^{(v)(v^{\prime})}=-\frac{1}{2 N} \sum_{n=1}^{N}\log \\&\frac{\exp \left(cos\left(t_{n}^{v}, t_{n}^{v^{\prime}}\right) / \tau_{t}\right)}{\sum_{m=1}^{N}\left[\exp \left(cos\left(t_{n}^{v}, t_{m}^{v}\right) / \tau_{t}\right)+\exp \left(cos\left(t_{n}^{v}, t_{m}^{v^{\prime}}\right) / \tau_{t}\right)\right]},
\end{aligned}
\end{equation}
where $\tau_{t}$ is the temperature parameter. Considering all views on the dataset, we give the optimization objective of the teacher network as
\begin{equation}
\label{eq8}
\begin{aligned}
\mathcal{L}_{tea}=\sum_{v=1}^{V} \sum_{v \neq v^{\prime}} \ell_{t}^{(v)(v^{\prime})}.
\end{aligned}
\end{equation}

\subsubsection{Invariant Information Clustering (IIC) Loss}
\begin{figure}[htbp]
\centering
\includegraphics[width=0.45\linewidth]{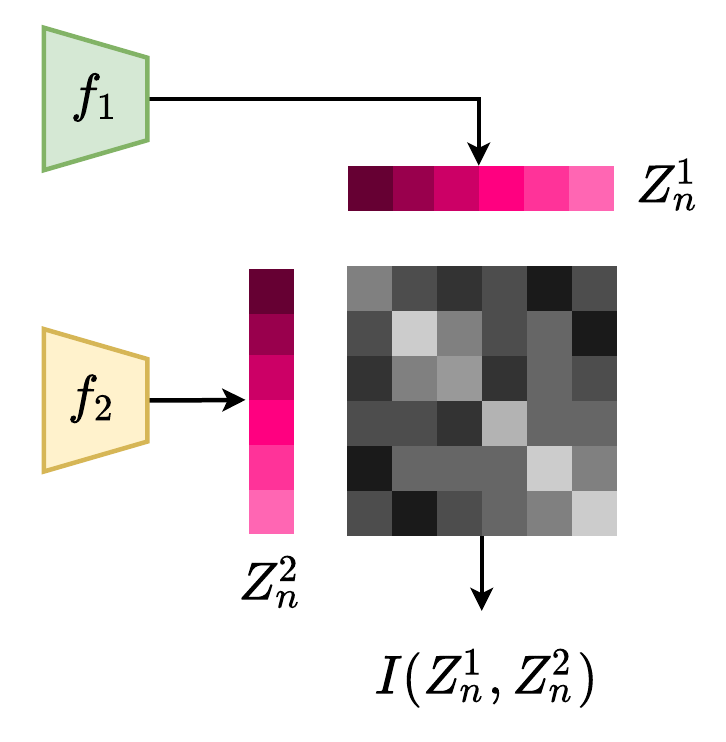}
\caption{Calculation of mutual information between two views. The mutual information of $Z^{1}_{n}$ and $Z^{2}_{n}$ can be directly obtained on a joint probability distribution matrix $P_{Z_n^1,Z_n^2}$. The matrix can be calculated by approximating $Z^{1}_{n}$ and $Z^{2}_{n}$ as two independent discrete probability distributions.}
\label{fig4}
\end{figure}
Minimizing InfoNCE \cite{34} at high-dimensional hierarchies can be seen as maximizing the lower bound of mutual information indirectly. That is, $I\left(y^{v}, y^{v^{\prime}}\right) \geq \log (n^{2}-n)-\mathcal{L}_{\mathrm{stu}}$, where $I\left(y^{v}, y^{v^{\prime}}\right)$ denotes the mutual information between $s^{v}$ and $s^{v^{\prime}}$, $(n^{2}-n)$ is the number of negative samples, and similarly $I\left(t^{v}, t^{v^{\prime}}\right) \geq \log (n^{2}-n)-\mathcal{L}_{\mathrm{tea}}$. Different from the above methods, we directly maximize the mutual information between different views in low-dimensional hierarchies: 
\begin{equation}
\label{eq9}
\begin{aligned}
\mathcal{L}_{IIC}=-\sum_{v=1}^{V} \sum_{v \neq v^{\prime}} \sum_{n=1}^{N} I\left(Z_{n}^{v}, Z_{n}^{v^{\prime}}\right),
\end{aligned}
\end{equation}
where $I$ represents mutual information. As shown in Figure 3, according to invariant information clustering (IIC) \cite{23}, we approximate $Z_{n}^{v}$ and $Z_{n}^{v^{\prime}}$ into two independent discrete distributions and further obtain the joint probability distribution of $Z_{n}^{v}$ and $Z_{n}^{v^{\prime}}$. Therefore, $I$ is directly calculated by
\begin{equation}
\label{eq10}
\begin{aligned}
\mathbb{E}_{P_{Z_{n}^{1}, Z_{n}^{2}}}\left(P_{Z_{n}^{1}, Z_{n}^{2}} \log \frac{P_{Z_{n}^{1}, Z_{n}^{2}}}{P_{Z_{n}^{1}}P_{Z_{n}^{2}}}\right).
\end{aligned}
\end{equation}

\subsection{Self-distillnation Loss}
\label{3.5}
To make better use of the learned common semantics for clustering, we need to add some interactions for the two independent student and teacher subspaces for fine-tuning. The teacher network and the student network use the same network structure, but the network parameters are different. The teacher network is updated in the form of a moving average \cite{33}, introducing a momentum encoder to provide a regression target for the student network. The parameter $\theta$ is $\xi$ exponential moving average. With the target momentum being $\mu \in[0,1]$, the parameter $\theta$ is updated with:

\begin{equation}
\label{eq11}
\begin{aligned}
\theta \leftarrow \mu \theta+(1-\mu) \xi.
\end{aligned}
\end{equation}

We do not use the soft labels output by the teacher network directly as the distribution required for distillation because such probability distributions do not contain obvious clustering information. We will first use the cluster information contained in the high-level features to improve the clustering effect of semantic labels, and a new cluster center $C$ can be obtained by optimizing the following objectives:
\begin{equation}
\label{eq12}
\begin{aligned}
\begin{aligned}\mathcal{L}_{Km} &=\min _{\left\{\mathbf{C}^{v}\right\}_{v=1}^{V}} \sum_{n \in \mathcal{X}} \sum_{m=1}^{K} \sum_{v=1}^{V}\left\|\theta z_{n}^{v}-c_{m}^{v}\right\|_{2}^{2} \\&=\min _{\mathbf{C}} \sum_{n \in \mathcal{X}} \sum_{m=1}^{K}\left\|t_{n}-c_{m}\right\|_{2}^{2}\end{aligned}.
\end{aligned}
\end{equation}
where $\theta$ is the parameter of the teacher network, $\mathbf{C} \in \mathbb{R}^{K \times \sum_{v=1}^{V} d_{v}}$, $c_{m}=\left(c_{m}^{1}, c_{m}^{2}, \ldots, c_{m}^{V}\right) \in \mathbb{R}^{K \times \sum_{v=1}^{V} d_{v}}$, and $d_{v}$ is the dimension of $t_{n}$. This step is more efficient with the K-means algorithm, so we can linearly separate the $t_{n}$ according to the cluster center $c$ to get the $V$ group of pseudo-labels $\left\{\mathbf{P}^{v} = {\operatorname{argmin}_{m}}\left\|t_{n}^{v}-c_{n}^{v}\right\|_{2}^{2}\right\}_{v=1}^{V}$. The Softmax activation function will be stacked to the predictor's final layer, and $s_{nm}^{v}$ is defined as the probability that the $n$-th sample is clustered into the $m$-th cluster for the $v$-th view, so there are also $V$ groups of probability distributions $\left\{\mathbf{l}^{v} = \operatorname{argmax}_{m} y_{n m}^{(v)}\right\}_{v=1}^{V}$. However, $\mathbf{P}^{v}$ and $\mathbf{I}^{v}$ are not aligned, so we need to define a loss matrix $\mathbf{M} \in \mathbb{R}^{K \times K}$ to help us correct $\mathbf{P}^{v}$ \cite{47}, $\tilde{\mathbf{m}}_{nm}=\sum_{n \in \mathcal{X}} \mathbb{1}\left[l_{i}^{v}=n\right] \mathbb{1}\left[l_{i}^{v}=m\right]$, element $\mathbf{m}_{n m}=\max _{n, m} \tilde{\mathbf{m}}_{n m}-\tilde{\mathbf{m}}_{n m}$. The alignment problem will be treated as a maximum matching problem:
\begin{equation}
\label{eq13}
\begin{aligned}
\min _{\mathbf{A}} \sum_{i=1}^{K} \sum_{j=1}^{K} m_{i j} a_{i j}\\\text { s.t. } \mathbf{A} \mathbf{A}^{T}=\mathbf{I}_{K},
\end{aligned}
\end{equation}
where $\mathbf{A} \in \mathbb{R}^{K \times K}$ is a Boolean matrix, and Equation \ref{eq13} is optimized using the Hungarian algorithm \cite{48}. With $\left\{\mathbf{P}^{*v}\right\}_{v=1}^{V}$ being the obtained dark knowledge,  we use the KL divergence distillation model:
\begin{equation}
\label{eq14}
\begin{aligned}
\mathcal{L}_{self}=-\sum_{v=1}^{V} [(1-\tau_{d} )\mathbf{P}^{*v}+ \tau_{d} u]\log \frac{[(1-\tau_{d})\mathbf{P}^{*v}+ \tau_{d} u]}{y^{v}},
\end{aligned}
\end{equation}
where $\tau_{d}$ is a distillation factor, $u$ is a distribution introduced, here is a Gaussian distribution. The softmax function makes $y^{v}$ a relatively sharp distribution, and dark knowledge is a relatively smooth distribution, and KL divergence can make the two form a confrontation, thereby effectively preventing the model from collapsing. Empirically, we set $\tau_{d}=0.1$. 

\subsection{Training and Inference}
$\mathcal{L}_{r e c}$ is the reconstruction loss of the autoencoder, $\mathcal{L}_{con}$ and $\mathcal{L}_{self}$ implement feature learning and label distillation, respectively. A dynamic balance factor is usually used to measure the loss throughout the training process \cite{35}. But in practice, we have found that simply adding together all these losses works well, so there is no need to set the balance factor. 

During the pretraining stage, we fed the dataset $\mathcal{X}$ to DistilMVC and use $(\mathcal{L}_{r e c}+\mathcal{L}_{c o n})$ as the objective function for training. Learning different hierarchies of mutual information can provide rich semantic knowledge, which lays the foundation for subsequent distillation. The pre-trained model is loaded and fine-tuned by optimizing $\mathcal{L}_{self}$ to alleviate the wrong traction of pseudo-labels and improve the clustering performance.

In the inference stage, we fed the entire dataset to DistilMVC, and the predictor $w_{p}$ in the student network branch will obtain the probability distribution of all view clusters $\left\{y_{n m}^{(v)}\right\}_{v= 1}^{V}$, the probability is weighted and summed on each view to get the final clustering result ${\operatorname{argmax}_{m}}\left(\frac{1}{V} \ sum_{v=1}^{V} y_{n m}^{v}\right)$.
\section{EXPERIMENTS}\label{sec4}

In this section, we     evaluate the proposed DistilMVC method on eight widely-used multi-view datasets and compare it with eight state-of-the-art clustering methods.

\subsection{Datasets and Experimental Settings}

\subsubsection{Comparisons with State of the Arts }
The comparison methods include two traditional methods (i.e., MVC-LFA \cite{MVC-LFA}, and IMVTST-MVI\cite{IMVTSC-MVI}) and six deep methods (i.e., CDIMC-net \cite{22}, EAMC \cite{eamc}, SiMVC \cite{SiMVC}, CoMVC \cite{SiMVC}, COMPLETER \cite{21}, SURE\cite{62} and MFLVC \cite{MFLVC}). For all methods, we use the recommended model structure and parameters for fair comparisons.

\subsubsection{Datasets}

\begin{table*}[]
\renewcommand\arraystretch{0.9}
\centering
\caption{Dataset summary}
\label{table1}
\setlength{\tabcolsep}{2.1mm}{
\begin{tabular}{lllccl}
\hline
Datasets   & Sample & Type                             & Views & \# of categories & Dimension           \\ \hline
Caltech-2V & 1,400  & WM and CENTRIST                  & 2     & 7                & 40/254              \\
Scene      & 4,485  & PHOG and GIST                    & 2     & 15               & 20/59               \\
MNIST-USPS & 5,000  & Two styles of digital images     & 2     & 10               & 784/784/784         \\
BDGP       & 2,500  & Visual and textual views         & 2     & 5                & 1750/79             \\
Caltech-3V & 1,400  & WM, CENTRIST, and LBP            & 3     & 7                & 40/254/928          \\
Caltech-4V & 1,400  & WM, CENTRIST, LBP, and GIST      & 4     & 7                & 40/254/928/512      \\
Caltech-5V & 1,400  & WM, CENTRIST, LBP, GIST, and HOG & 5     & 7                & 40/254/928/512/1984 \\ 
Fashion    & 10,000 & Three styles of images \cite{50} & 3     & 10               & 784/784/784         \\ \hline
\end{tabular}
}
\end{table*}

In our experiments, we used eight datasets: Scene \cite{Scene}, MNIST-USPS \cite{MNIST-USPScomic}, BDGP \cite{bdgp}, Fashion \cite{fashion}, Caltech-2V, Caltech-3V, Caltech-4V, and Caltech-5V. To evaluate the robustness of DistilMVC over the number of views, Caltech \cite{Caltech} as a multi-view RGB image dataset is disassembled into Caltech-2V, Caltech-3V, Caltech-4V, and Caltech-5V. Table \ref{table1} describes the datasets used in more detail.

\subsubsection{Experimental implementation}
We conduct all the experiments on the platform of ubuntu 16.04 with Tesla P100 Graphics Processing Units (GPUs) and 32G memory size. Our model, method and baseline are built on the pytorch 1.11.0 framework. Based on extensive ablation studies, the bacth size is set to 128 and the epochs for the two phases of pretraining and fine-tuning were set to 150 and 50. The temperature parameters $\tau_{s}$, $\tau_{t}$ and $\tau_{d}$ are fixed to 0.5, 1.0 and 0.1, respectively. We use Adam optimizer \cite{adam} with the default parameters to train our model and set the initial learning rate as 0.0001. The structure of the autoencoder for the $v$-th view is defined as $X^{v}-Fc_{512}-Fc_{1024}-Fc_{2048}-Fc_{512}-Z^{v}-Fc_{512}-Fc_{2048}-Fc_{1024}-Fc_{512}-\hat{X}^{v}$, where $Fc_{512}$ denotes a fully connected neural network with 512 neurons, and each layer is followed by a ReLU layer. As shown in Fig. \ref{fig3}, the teacher network structure and the student network structure have two linear layers each, and the ReLU activation function is added in the middle.

\subsubsection{Evaluate Metrics}
The clustering performance is evaluated with three metrics: Accuracy (ACC), Normalized Mutual Information (NMI) and Purity (PUR). More details on these indicators can be found in \cite{49}. A higher value of these evaluation indicators can reflect a better clustering performance.

\subsection{Experimental Results and Analysis}

\begin{table*}[]
\caption{The performance comparisions on four dual-view datasets. The ${1^{\mathrm{st}}}$ best results are indicated in \color{red}red \color{black}and the ${2^{\text {nd }}}$ best results are indicated in \color{blue}blue.}
\label{table2}
\renewcommand\arraystretch{0.9}
\centering
\setlength{\tabcolsep}{1.8mm}{
\begin{tabular}{lcccccccccccc}
\hline
Datasets    & \multicolumn{3}{c}{Caltech-2V}    & \multicolumn{3}{c}{Scene}    & \multicolumn{3}{c}{MNIST-USPS}    & \multicolumn{3}{c}{BDGP}    \\
Evaluation metrics               & ACC      & NMI      & PUR      & ACC      & NMI      & PUR      & ACC      & NMI      & PUR      & ACC      & NMI      & PUR      \\ \hline
MVC-LFA \cite{MVC-LFA}(2019)     & 0.462    & 0.348    & 0.496    & 0.357    & 0.391    & 0.384    & 0.768    & 0.675    & 0.768    & 0.564    & 0.395    & 0.612    \\
CDIMC-net \cite{22}(2020)        & 0.515    & 0.480    & 0.564    & 0.346    & 0.374    & 0.351    & 0.620    & 0.676    & 0.647    & 0.884    & 0.799    & 0.885    \\
EAMC \cite{eamc}(2020)           & 0.419    & 0.256    & 0.427    & 0.250    & 0.319    & 0.263    & 0.735    & 0.837    & 0.778    & 0.681    & 0.480    & 0.697    \\
IMVTST-MVI\cite{IMVTSC-MVI}(2021)& 0.409    & 0.398    & 0.540    & 0.340    & 0.312    & 0.181    & 0.669    & 0.592    & 0.717    & 0.981    & 0.950    & 0.982    \\
SiMVC \cite{SiMVC}(2021)         & 0.508    & 0.471    & 0.557    & 0.289    & 0.281    & 0.293    & 0.981    & 0.962    & 0.981    & 0.704    & 0.545    & 0.723    \\
CoMVC \cite{SiMVC}(2021)         & 0.466    & 0.426    & 0.527    & 0.306    & 0.303    & 0.314    & 0.987    & 0.976    & 0.989    & 0.802    & 0.670    & 0.803    \\
COMPLETER \cite{21}(2021)        & 0.599    & \color{red}0.572    & 0.612    & 0.391    & 0.415    & 0.401    & 0.989    & 0.971    & 0.989    & 0.960    & 0.950    & 0.963  \\
SURE \cite{62}(2022)          & 0.548       & 0.471    & 0.580       & \color{blue}0.417   & 0.426        & 0.441   & 0.992    & 0.977    & 0.992    & 0.907   & 0.794    & 0.907   \\
MFLVC \cite{MFLVC}(2022)         & \color{blue}0.606   & 0.528    & \color{blue}0.616   & 0.401    & \color{blue}0.428    & \color{blue}0.443   & \color{blue}0.995   & \color{blue}0.985   & \color{blue}0.995   & \color{blue}0.989    & \color{blue}0.966   & \color{blue}0.989    \\
DistilMVC(ours)                  & \color{red}0.619    & \color{blue}0.533    & \color{red}0.619    & \color{red}0.428   & \color{red}0.432    & \color{red}0.448    & \color{red}0.996    & \color{red}0.987    & \color{red}0.996    & \color{red}0.991    & \color{red}0.971    & \color{red}0.991    \\ \hline
\end{tabular}
}
\end{table*}

\begin{table*}[]
\caption{The performance comparison over four multi-view datasets. The symbol ‘–’ denotes unknown results, as COMPLETER and SURE mainly focus on two-view clustering.}
\label{table3}
\renewcommand\arraystretch{0.9}
\centering
\setlength{\tabcolsep}{1.8mm}{
\begin{tabular}{lcccccccccccc}
\hline
Datasets    & \multicolumn{3}{c}{Caltech-3V}    & \multicolumn{3}{c}{Caltech-4V}    & \multicolumn{3}{c}{Caltech-5V}     & \multicolumn{3}{c}{Fashion}\\
Evaluation metrics                & ACC      & NMI      & PUR      & ACC      & NMI      & PUR      & ACC      & NMI      & PUR      & ACC      & NMI      & PUR       \\ \hline
MVC-LFA \cite{MVC-LFA}(2019)      & 0.551    & 0.423    & 0.578    & 0.609    & 0.522    & 0.636    & 0.741    & 0.601    & 0.747    & 0.791    & 0.759    & 0.794     \\
CDIMC-net \cite{22}(2020)         & 0.528    & 0.483    & 0.565    & 0.560    & 0.564    & 0.617    & 0.727    & 0.692    & 0.742    & 0.776    & 0.809    & 0.789     \\
EAMC \cite{eamc}(2020)            & 0.389    & 0.214    & 0.398    & 0.356    & 0.205    & 0.370    & 0.318    & 0.173    & 0.342    & 0.614    & 0.608    & 0.638     \\
IMVTST-MVI\cite{IMVTSC-MVI}(2021) & 0.558    & 0.445    & 0.576    & 0.687    & 0.610    & 0.719    & 0.760    & 0.691    & 0.785    & 0.632    & 0.648    & 0.635     \\
SiMVC \cite{SiMVC}(2021)          & 0.569    & 0.495    & 0.591    & 0.619    & 0.536    & 0.630    & 0.719    & 0.677    & 0.729    & 0.825    & 0.839    & 0.825     \\
CoMVC \cite{SiMVC}(2021)          & 0.541    & 0.504    & 0.584    & 0.568    & 0.569    & 0.646    & 0.700    & 0.687    & 0.746    & 0.857    & 0.864    & 0.863     \\
COMPLETER \cite{21}(2021)         & –        & –        & –        & –        & –        & –        & –        & –        & –        & –        & –        & –         \\
SURE\cite{62}(2022) & – & – & – & – & – & – & – & – & – & – & –  & –        \\
MFLVC \cite{MFLVC}(2022)        & \color{blue}0.631   & \color{blue}0.566  & \color{blue}0.639  & \color{blue}0.733    & \color{blue}0.652    & \color{blue}0.734    & \color{blue}0.804    & \color{blue}0.703    &\color{blue} 0.804   & \color{blue}0.992  & \color{blue}0.980  & \color{blue}0.992 \\
DistilMVC(ours)                 & \color{red}0.650    & \color{red}0.575   & \color{red}0.663   & \color{red}0.809     & \color{red}0.695     & \color{red}0.809     & \color{red}0.824    & \color{red}0.709     & \color{red}0.824    & \color{red}0.993   & \color{red}0.982   & \color{red}0.993 \\ \hline
\end{tabular}
}
\end{table*}

Table \ref{table2} and Table \ref{table3} list the clustering performances of all methods on eight datasets, from which we obtain the following observations: 
(1) Our DistilMVC achieves the best performance on all datasets. Compared with the second best method, DistilMVC has a significant improvement, especially surpassing 7.6\% on the Caltech-4V dataset. 
(2) COMPLETER and SURE suffer from the missing and unaligned data problems, respectively, so we evaluated the above two methods using complete and aligned data and found that they still significantly underperformed DistilMVC.
(3) PUR calculates the proportion of the samples in a cluster with the ground-true label\cite{49}. ACC only concerns about the best matched cluster with the ground-true label\cite{48}. 
Therefore, the case that some clusters share the same label will lead to PUR$>$ACC \cite{manning2008introduction}. 
Our DistilMVC obtains the same value for both ACC and PUR on all six datasets, which indicates that there is a strict one-to-one relation between the predicted clusters by DistilMVC and the ground-true clusters, i.e., no cluster's labels is duplicated, ensuring that the semantics of each predicted cluster are independent of each other (see Fig.~\ref{fig5}). In contrast, PUR values of all other methods are higher than their ACC values. This also confirms the robustness of our method.

\begin{figure*}[h!]
\centering
\subfigure[Caltech-2V]{
\includegraphics[width=3.6cm]{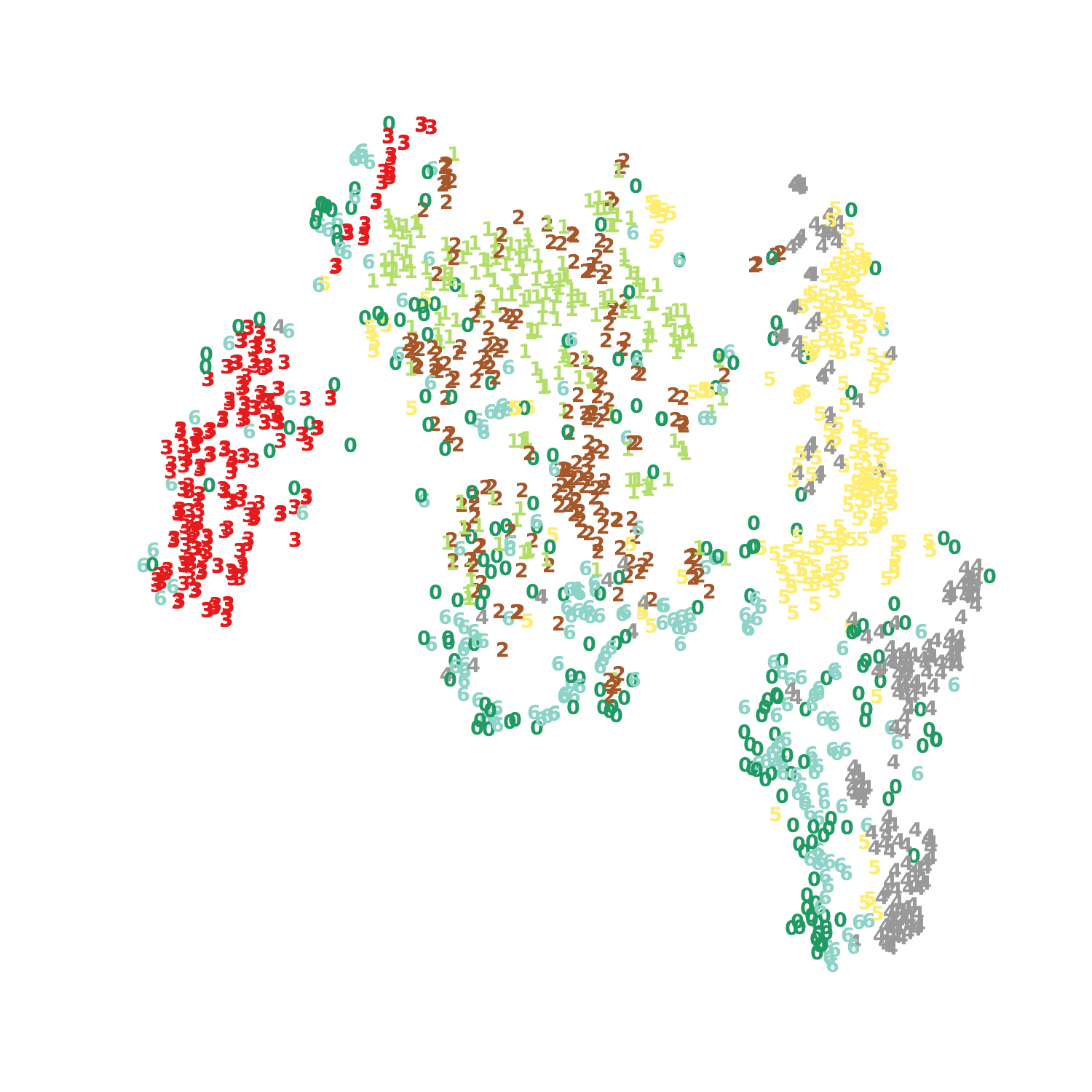}
\includegraphics[width=3.6cm]{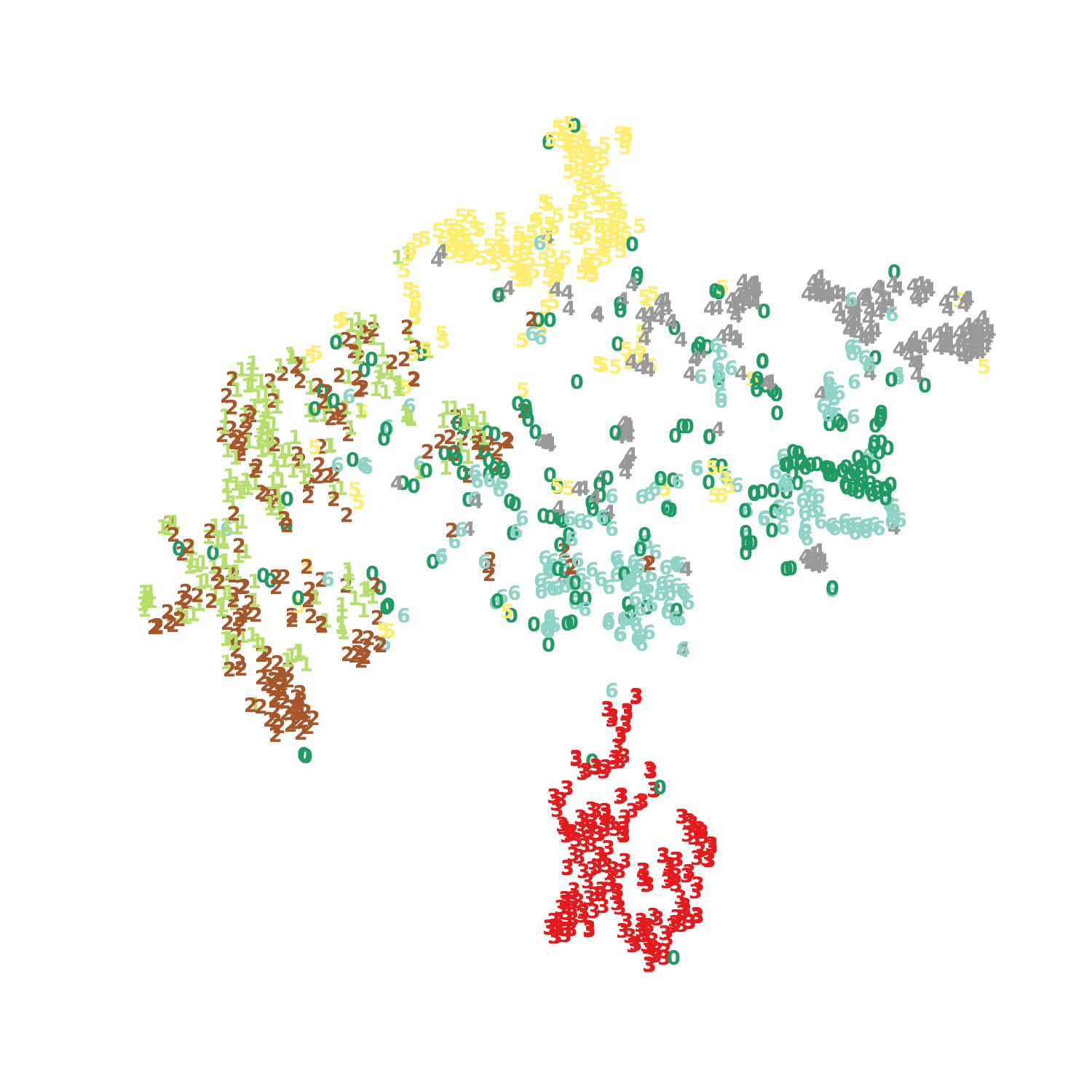}
}
\subfigure[Scene]{
\includegraphics[width=3.6cm]{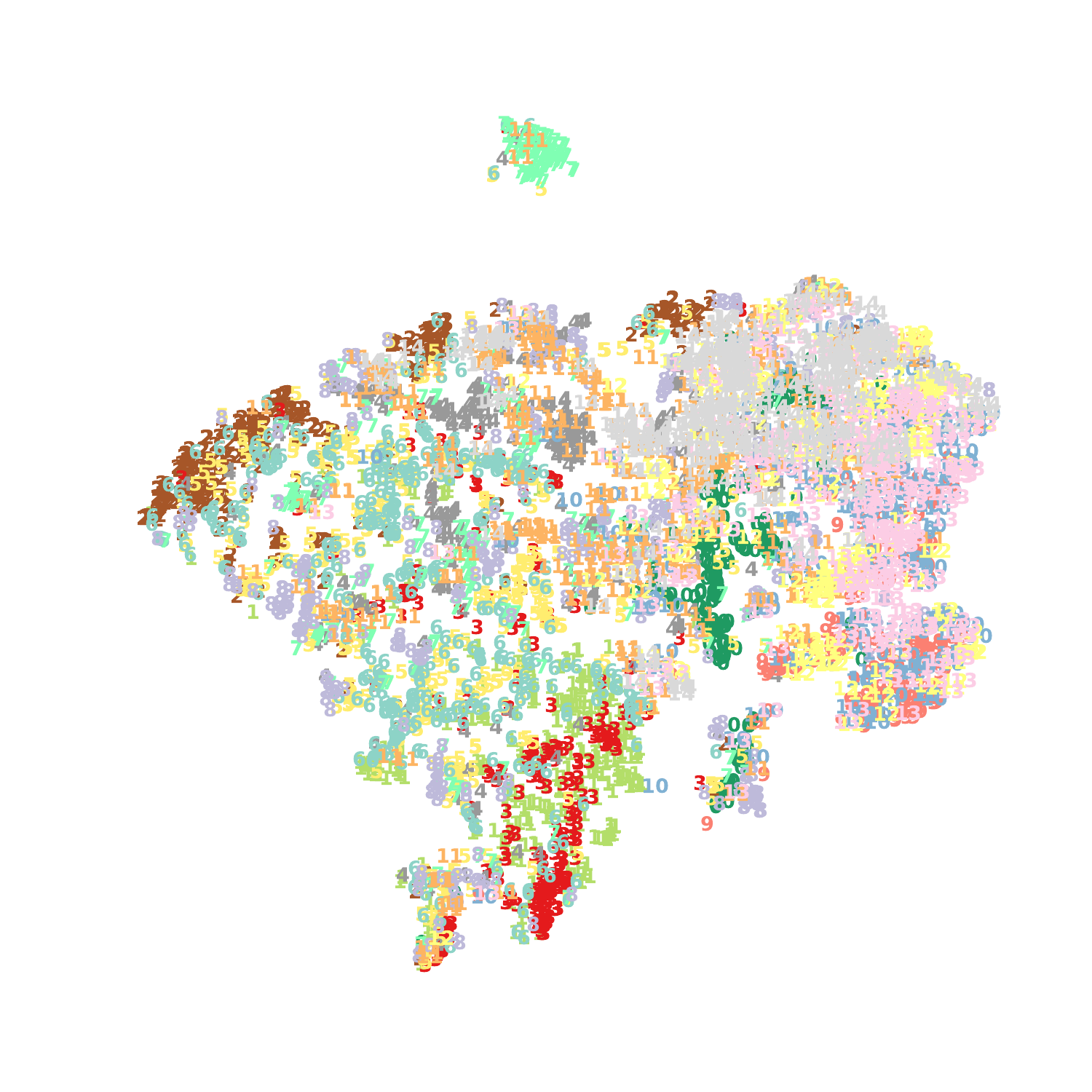}
\includegraphics[width=3.6cm]{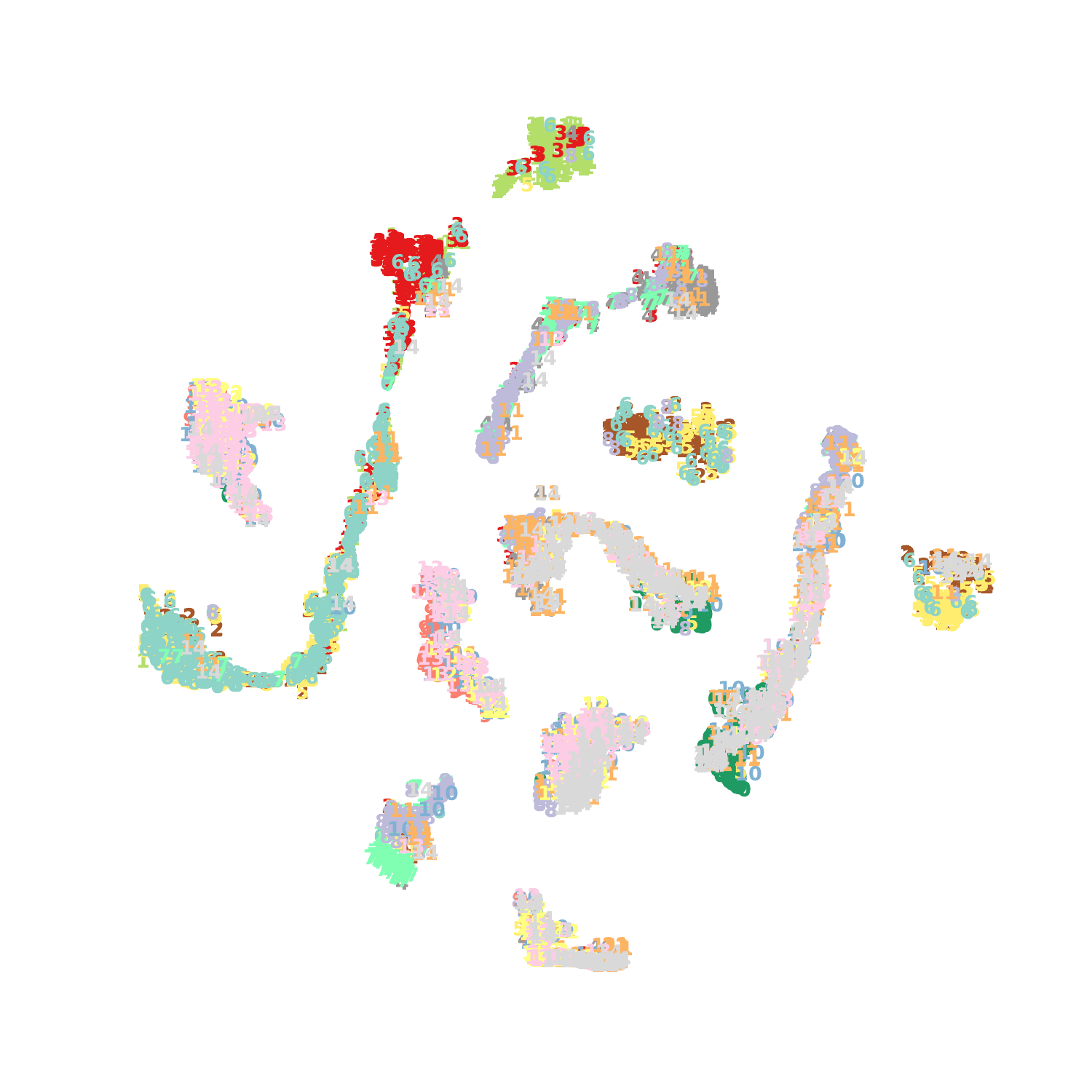}
}
\subfigure[MNIST-USPS]{
\includegraphics[width=3.6cm]{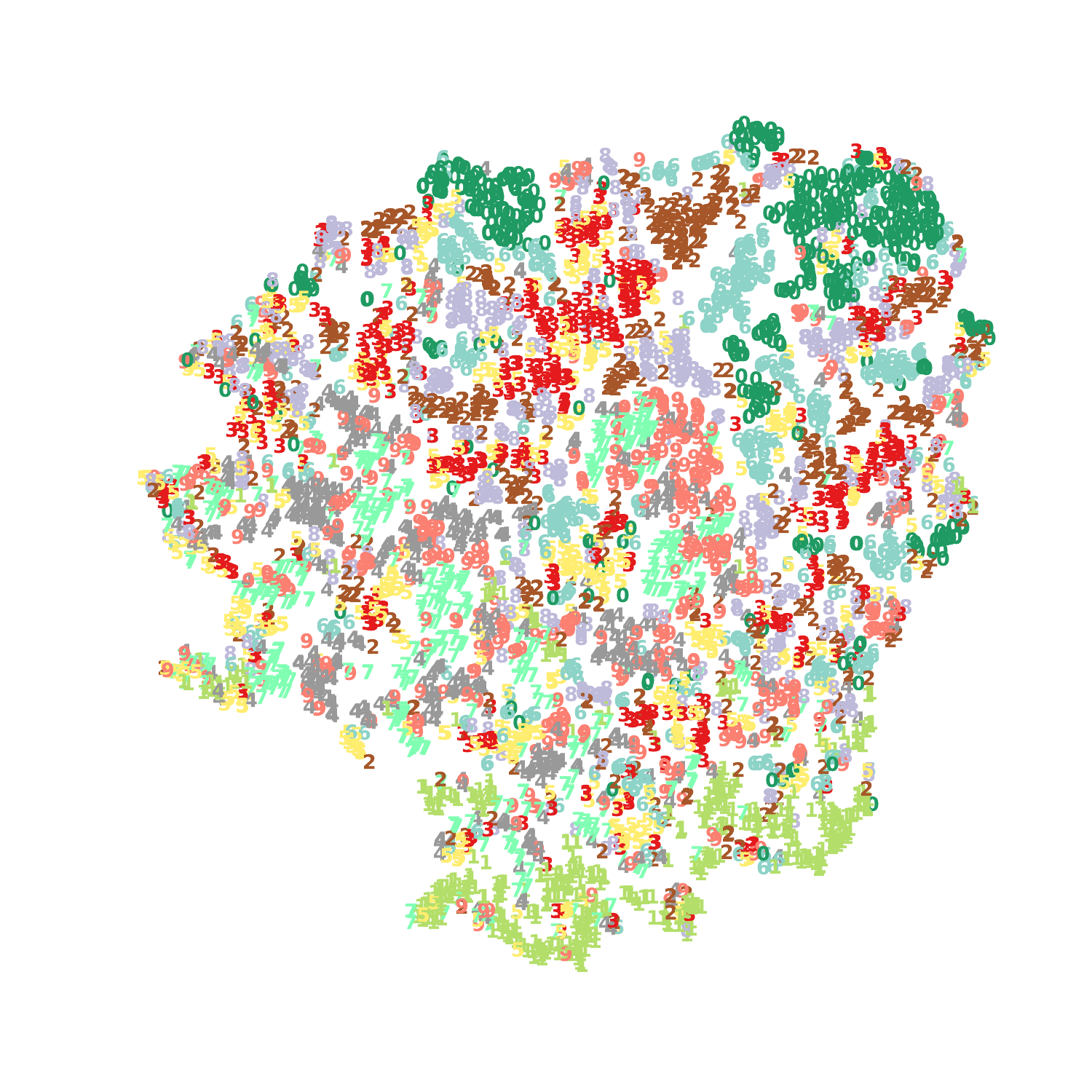}
\includegraphics[width=3.6cm]{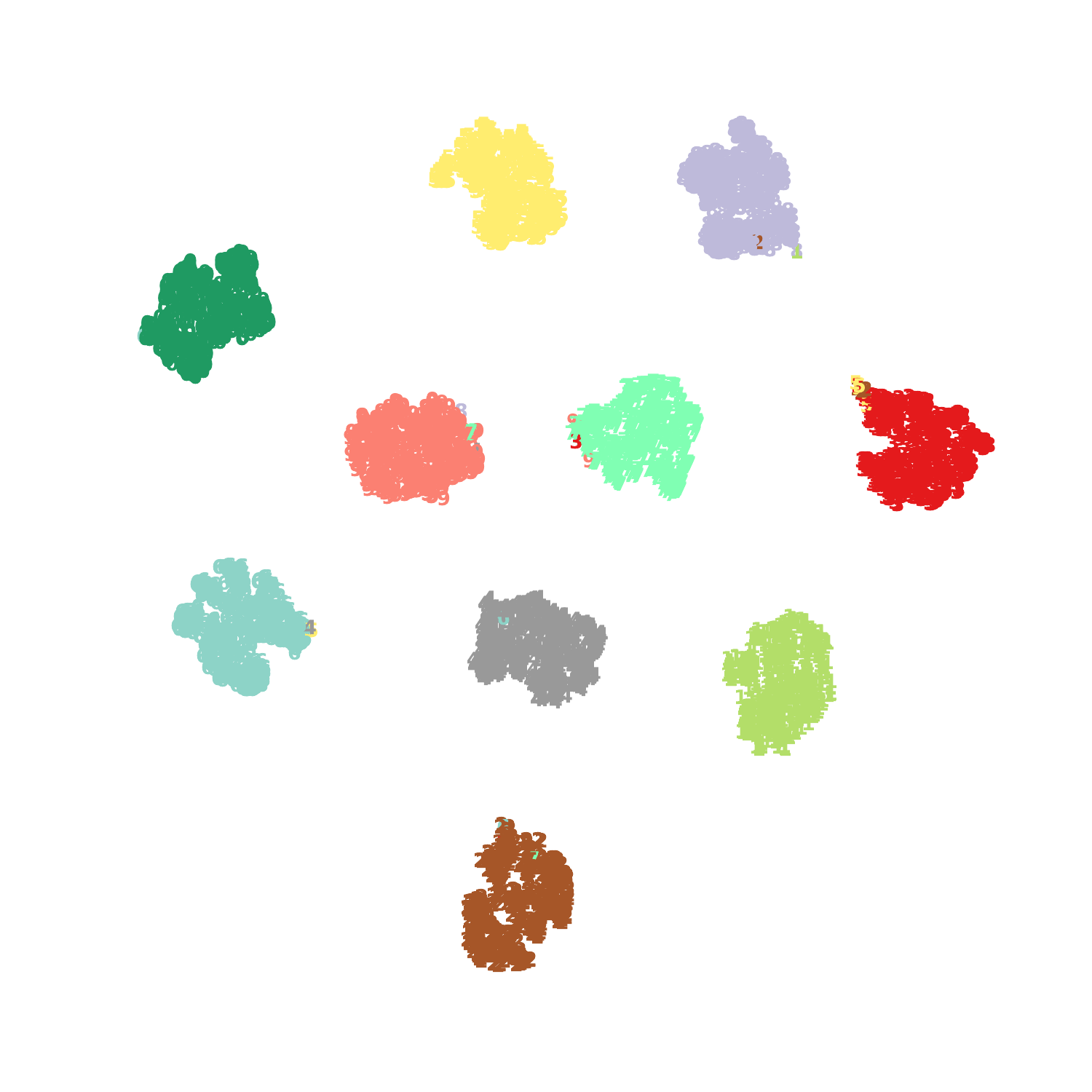}
}
\subfigure[BDGP]{
\includegraphics[width=3.6cm]{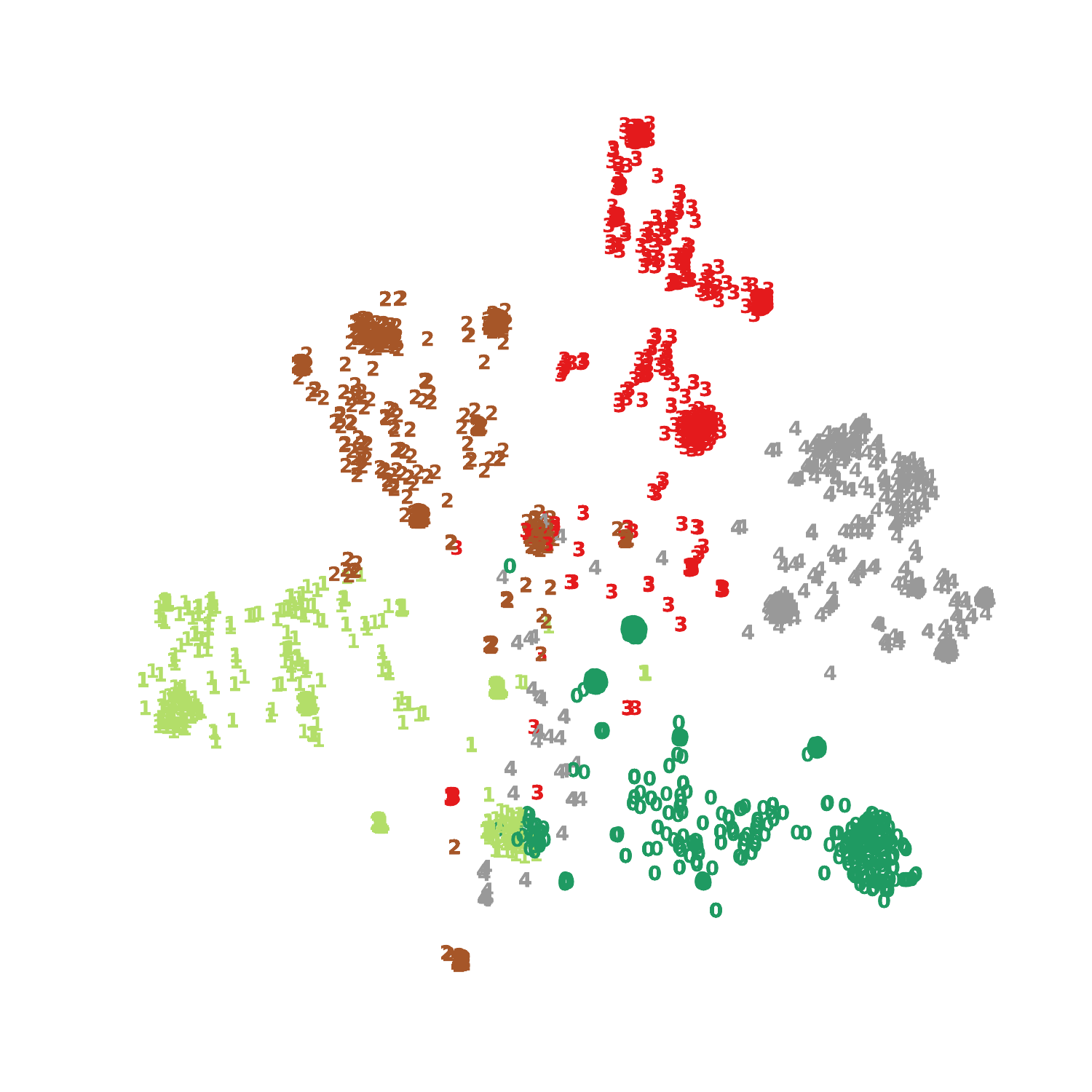}
\includegraphics[width=3.6cm]{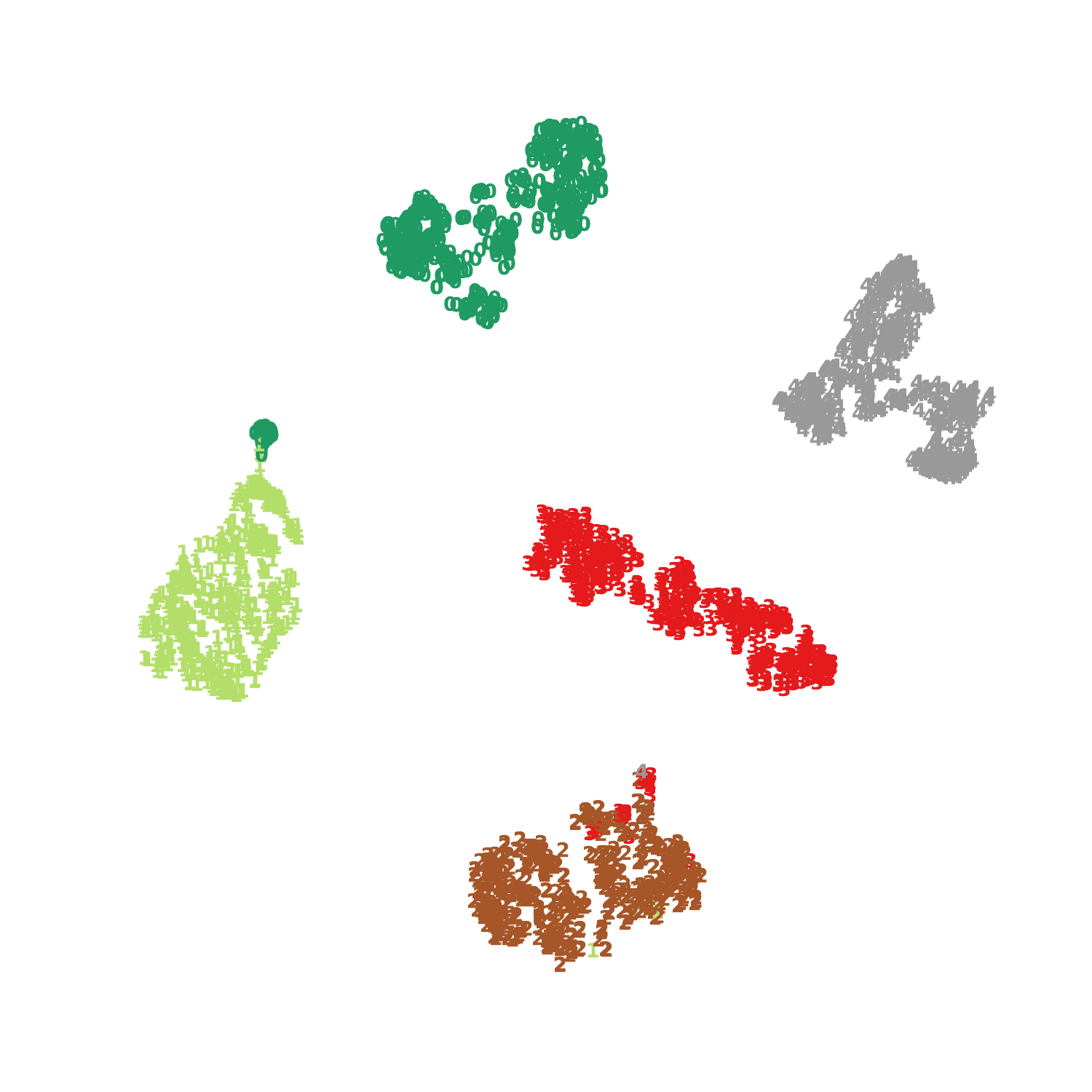}
}
\subfigure[Caltech-3V]{
\includegraphics[width=3.6cm]{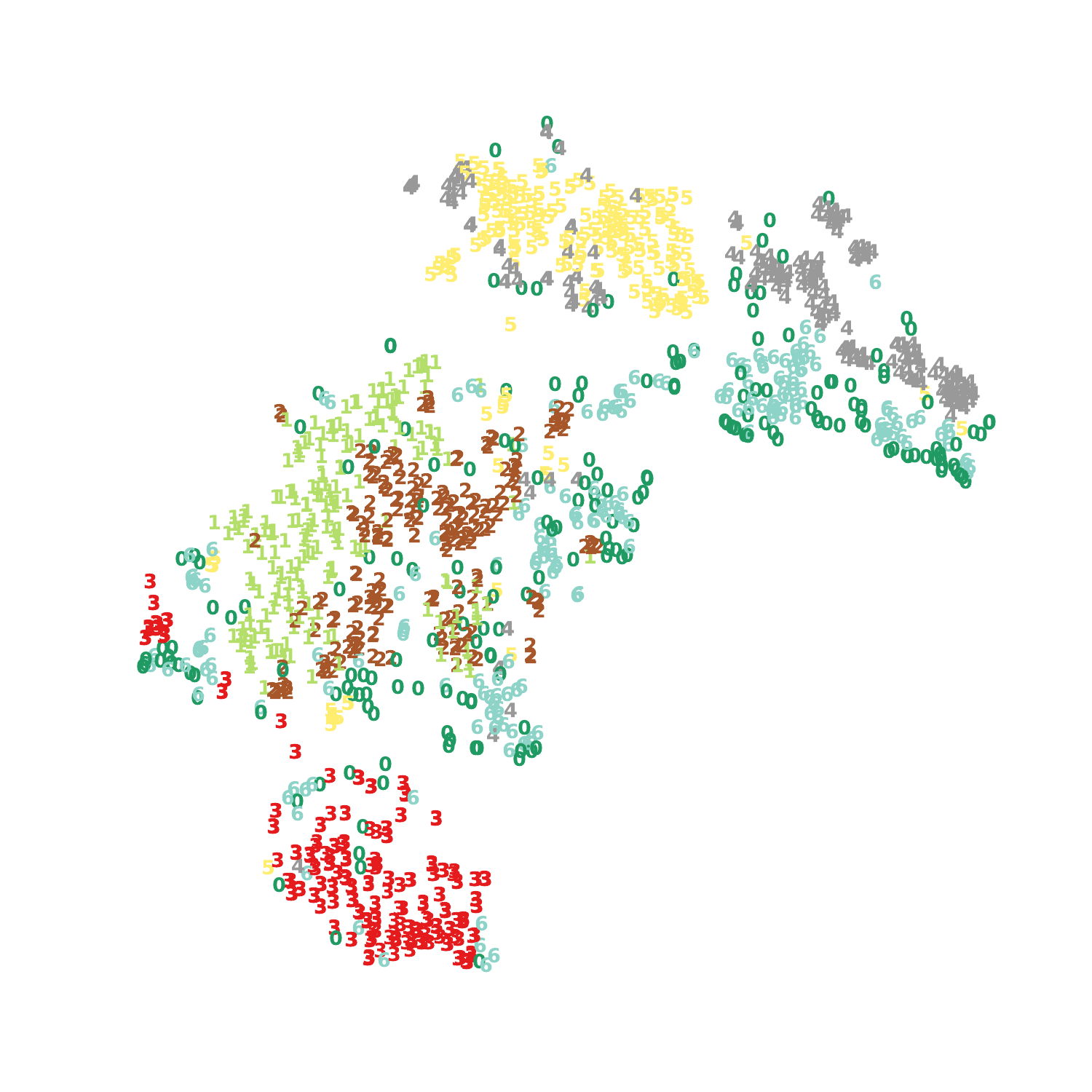}
\includegraphics[width=3.6cm]{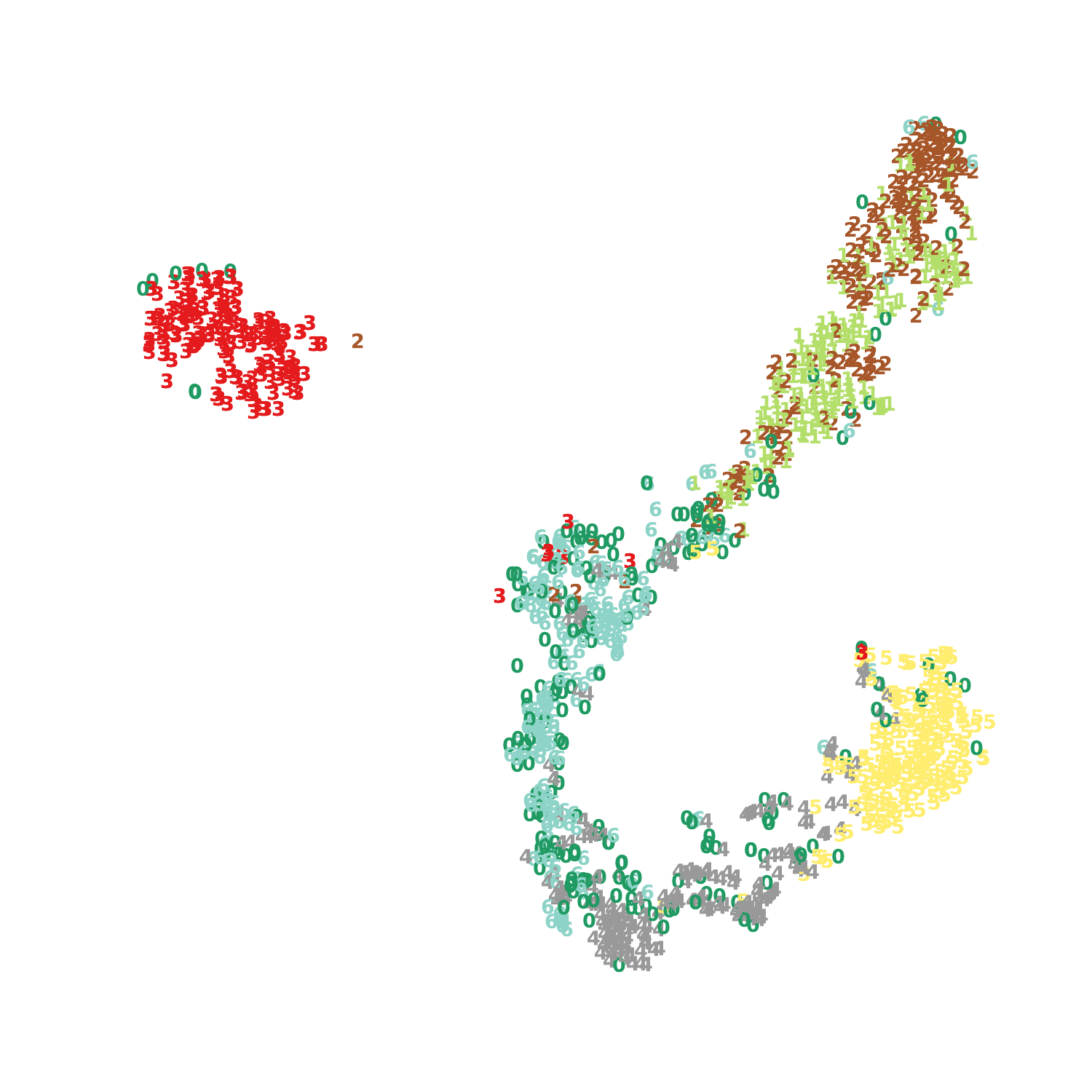}
}
\subfigure[Caltech-4V]{
\includegraphics[width=3.6cm]{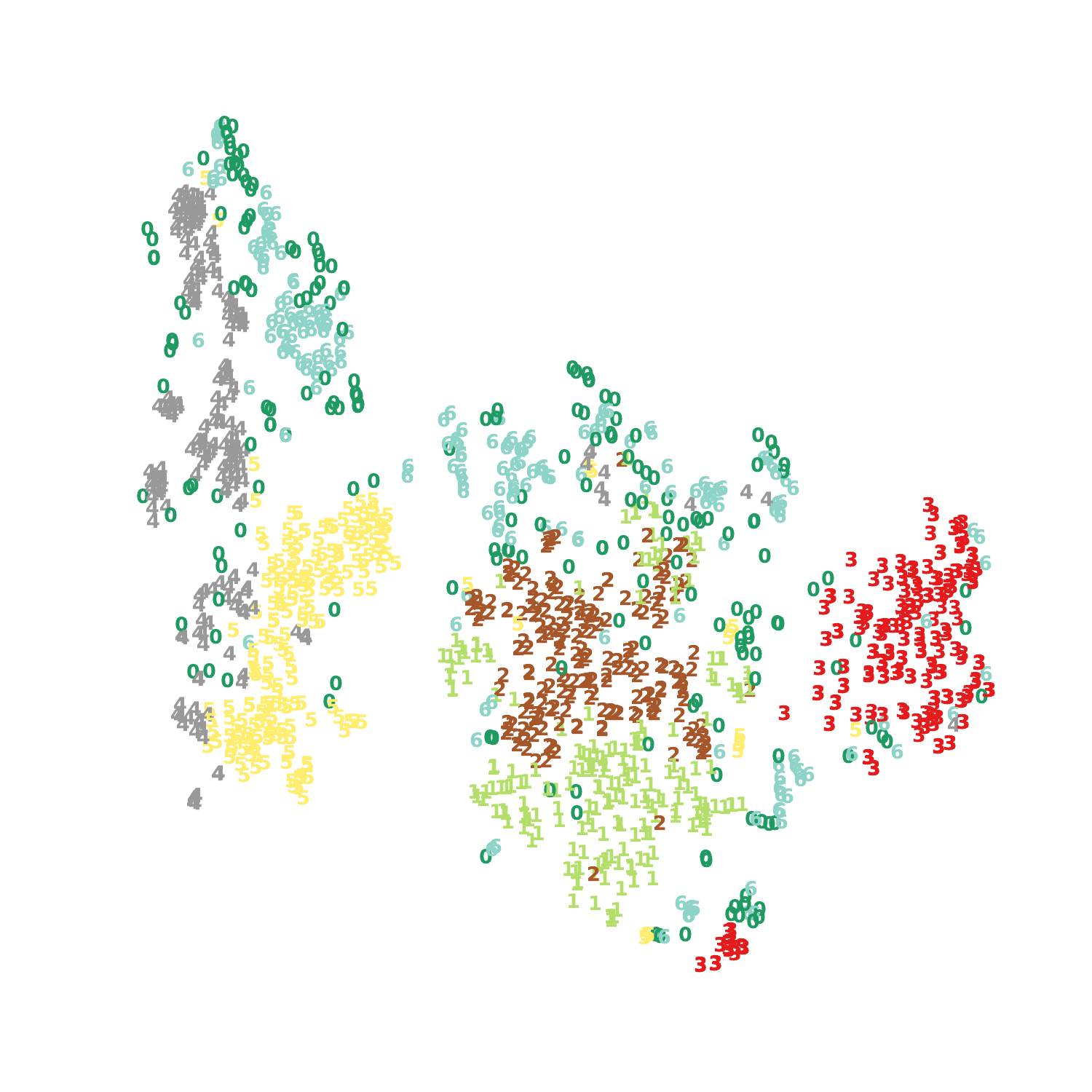}
\includegraphics[width=3.6cm]{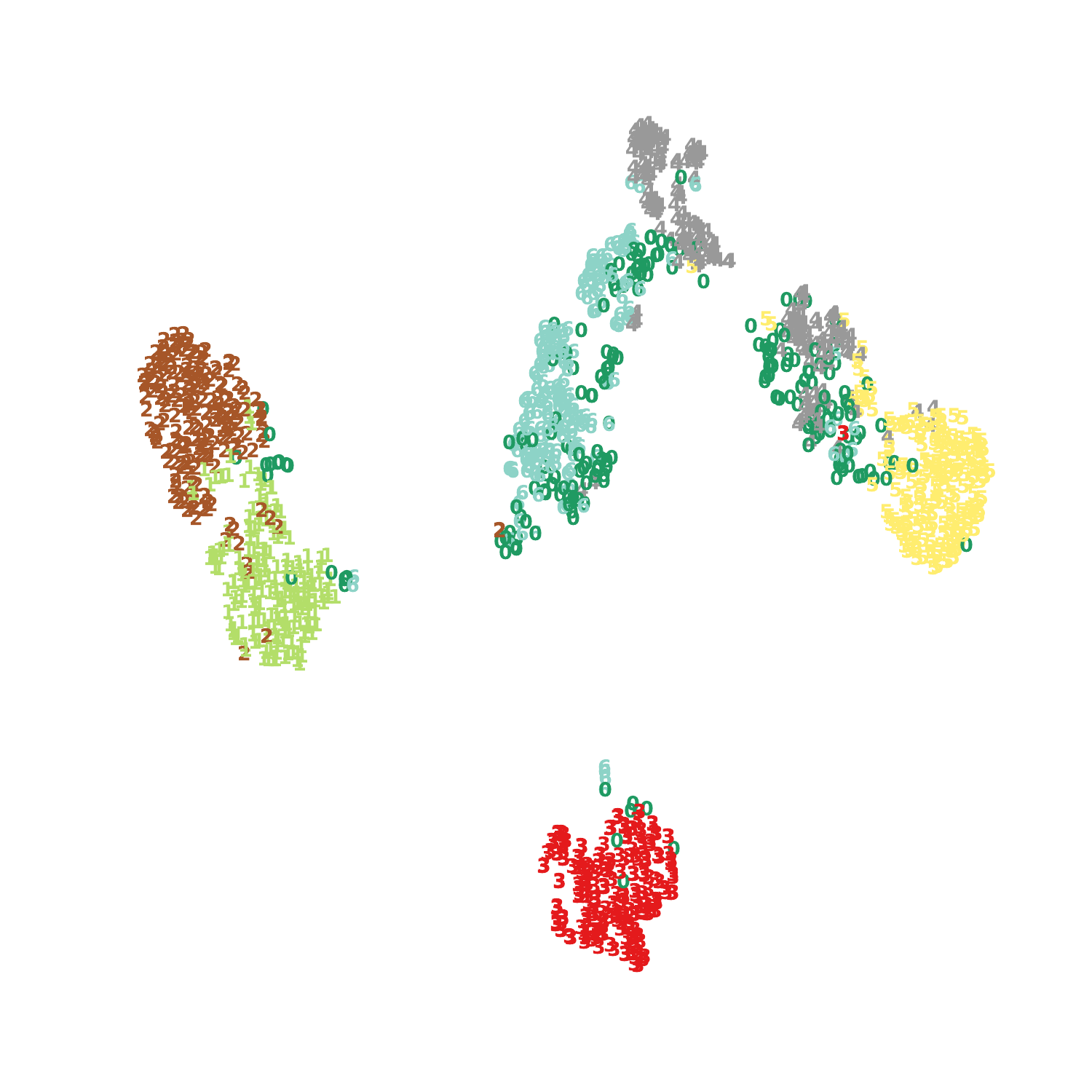}
}
\subfigure[Caltech-5V]{
\includegraphics[width=3.6cm]{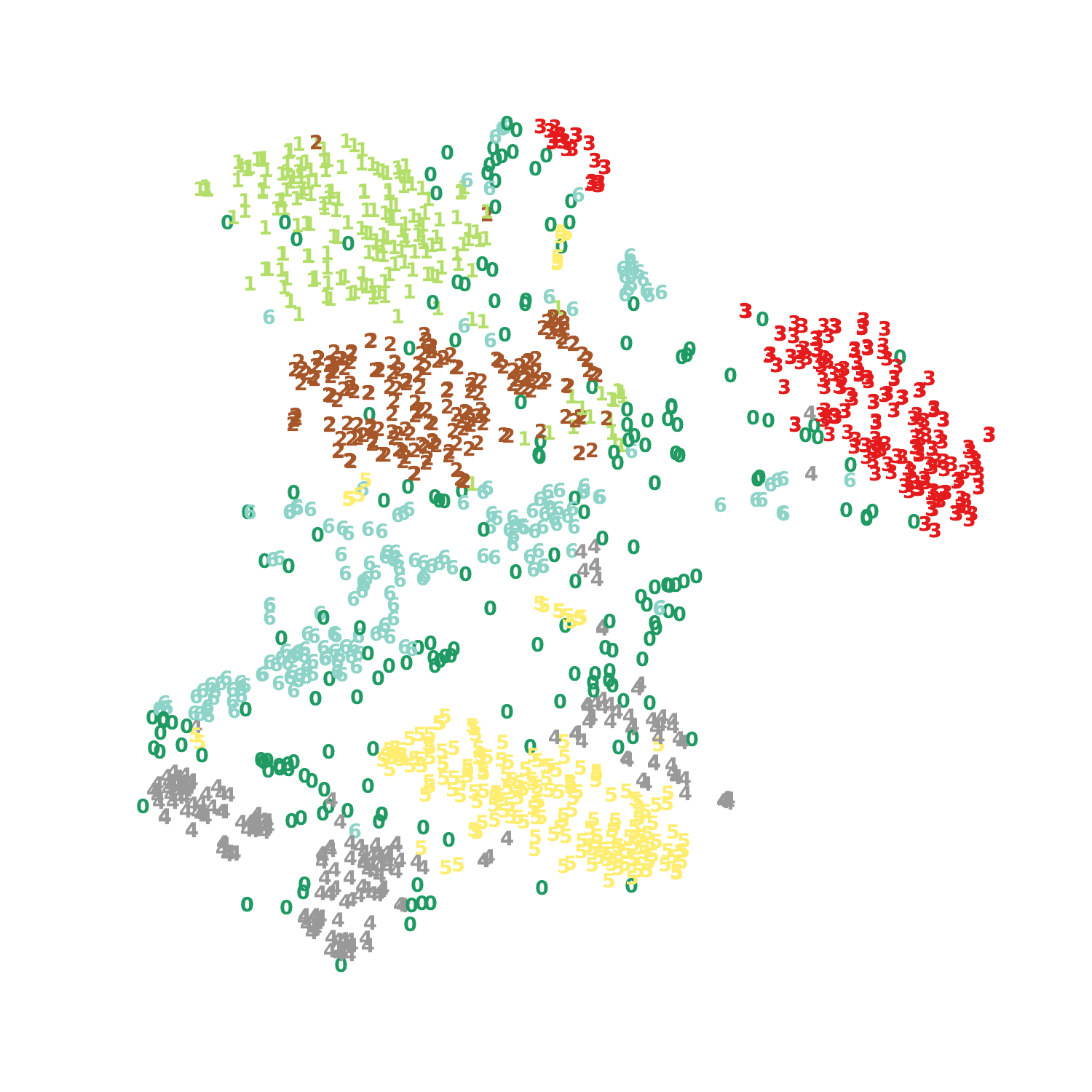}
\includegraphics[width=3.6cm]{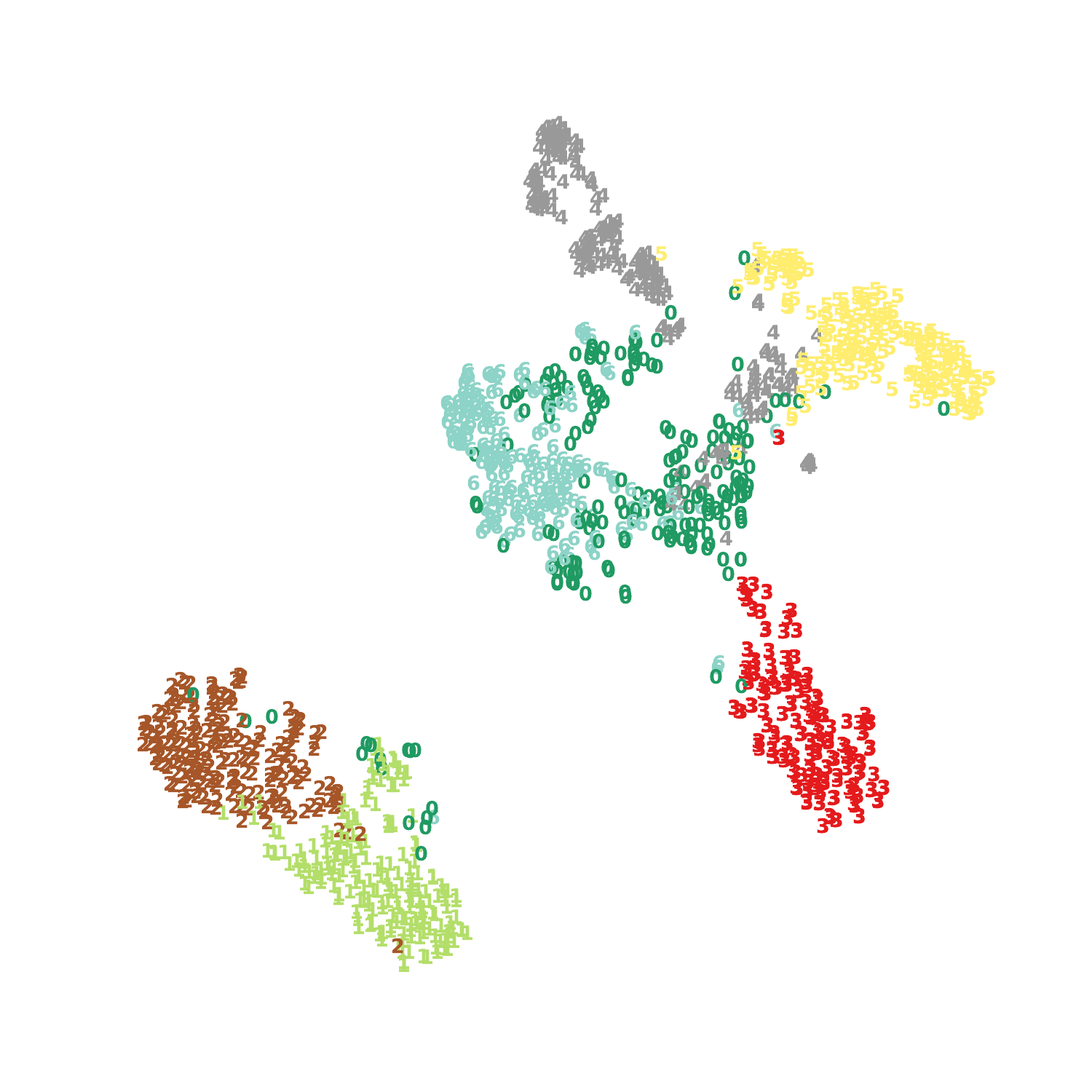}
}
\centering
\subfigure[Fashion]{
\includegraphics[width=3.6cm]{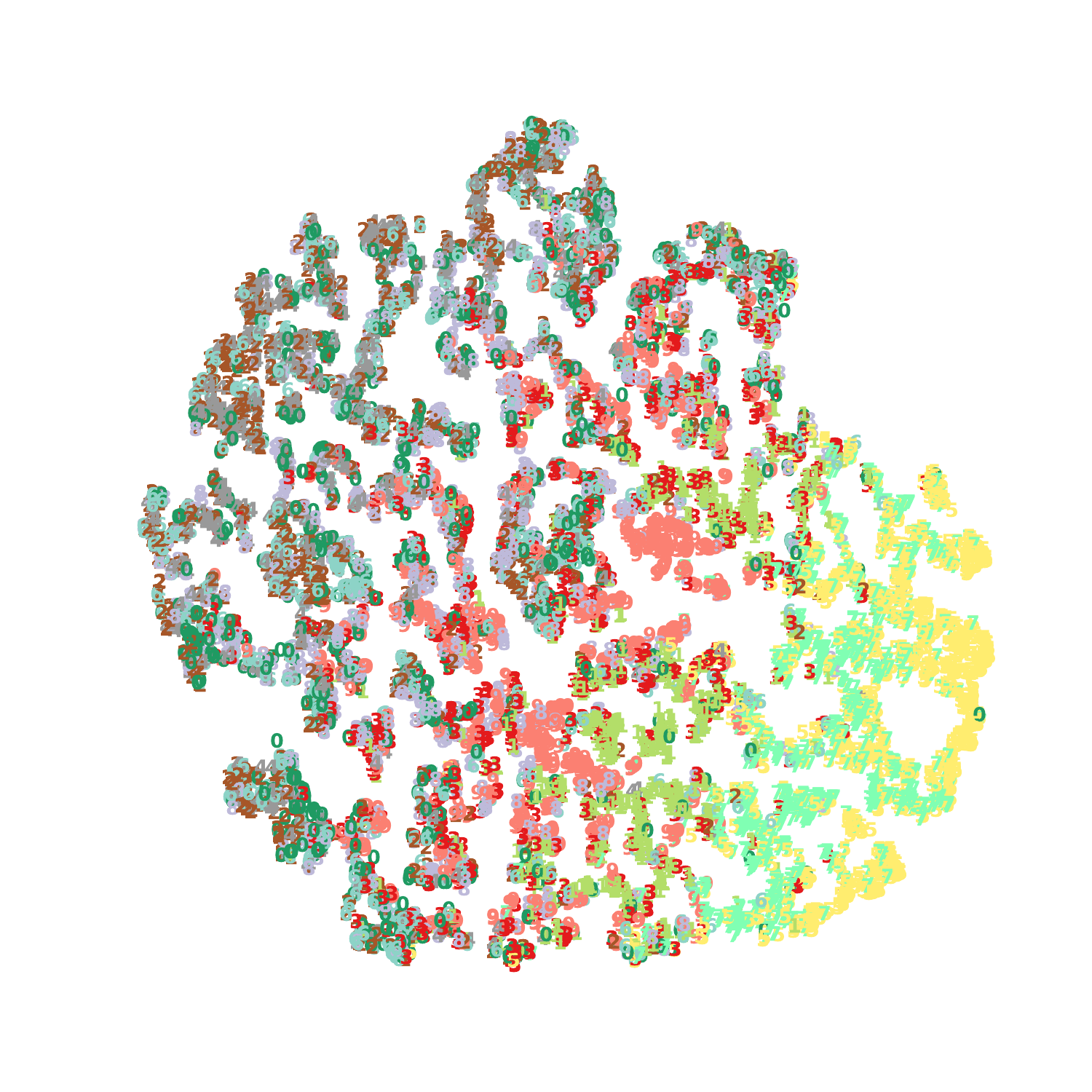}
\includegraphics[width=3.6cm]{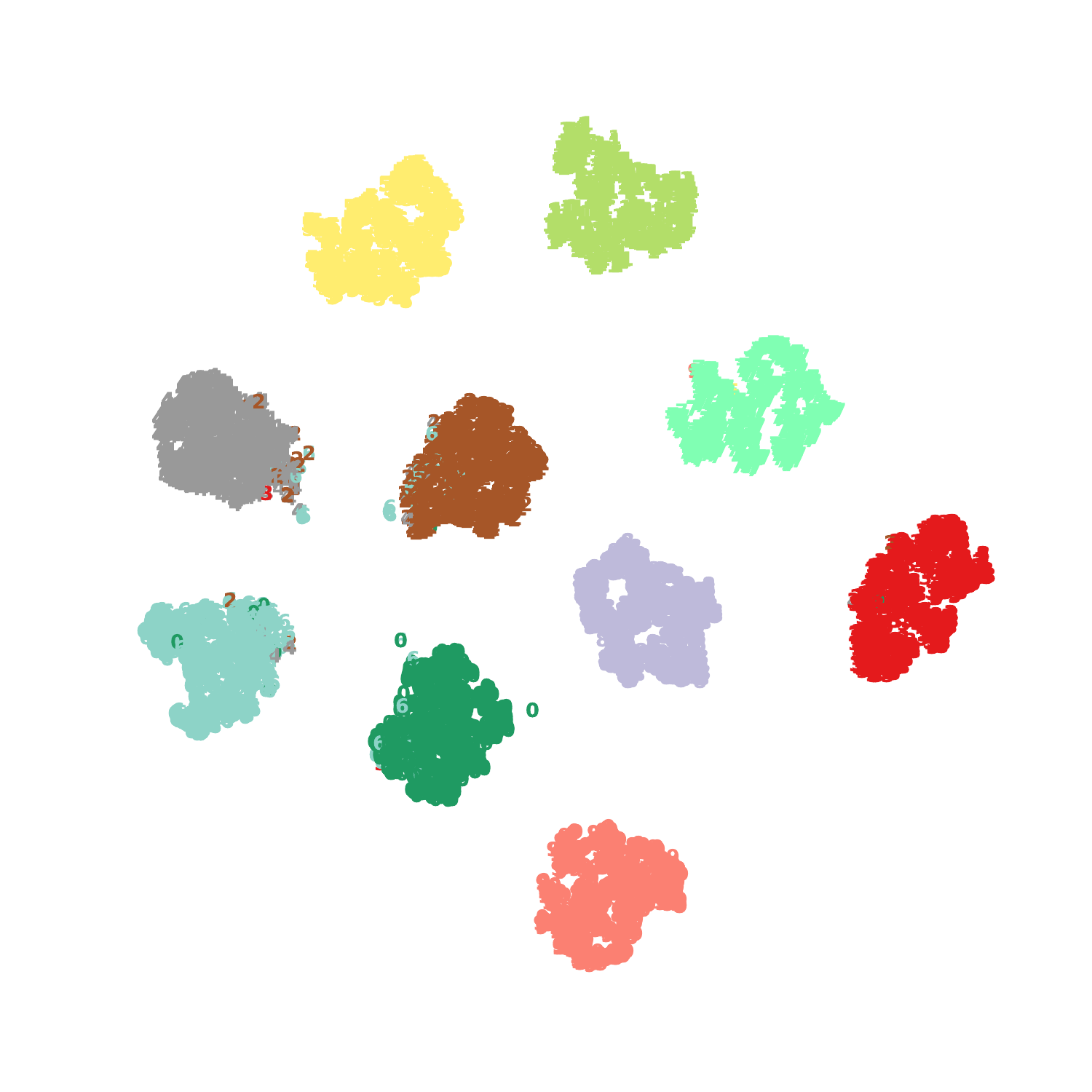}
}

\caption{Visualization on eight datasets via t-SNE \cite{51}. For each dataset, we visualize the fused representation of the different views and the fused representation obtained by the student network after DistilMVC training.}
\label{fig6}
\end{figure*}
The reasons for the above observations can be explained as follows: 
(1) None of the baselines take into account the over-confident traction of inaccurate pseudo-labels, resulting in limited clustering quality. 
(2) COMPLETER and SURE suffer from lacks in deep mining of mutual information at different hierarchies.
(3) PUR values of all other methods are higher than their ACC values, which means different predicted clusters shared the same label.

\begin{figure*}[htbp]
\centering
\subfigure[Convergence on Caltech-2V]{
\includegraphics[width=3.9cm]{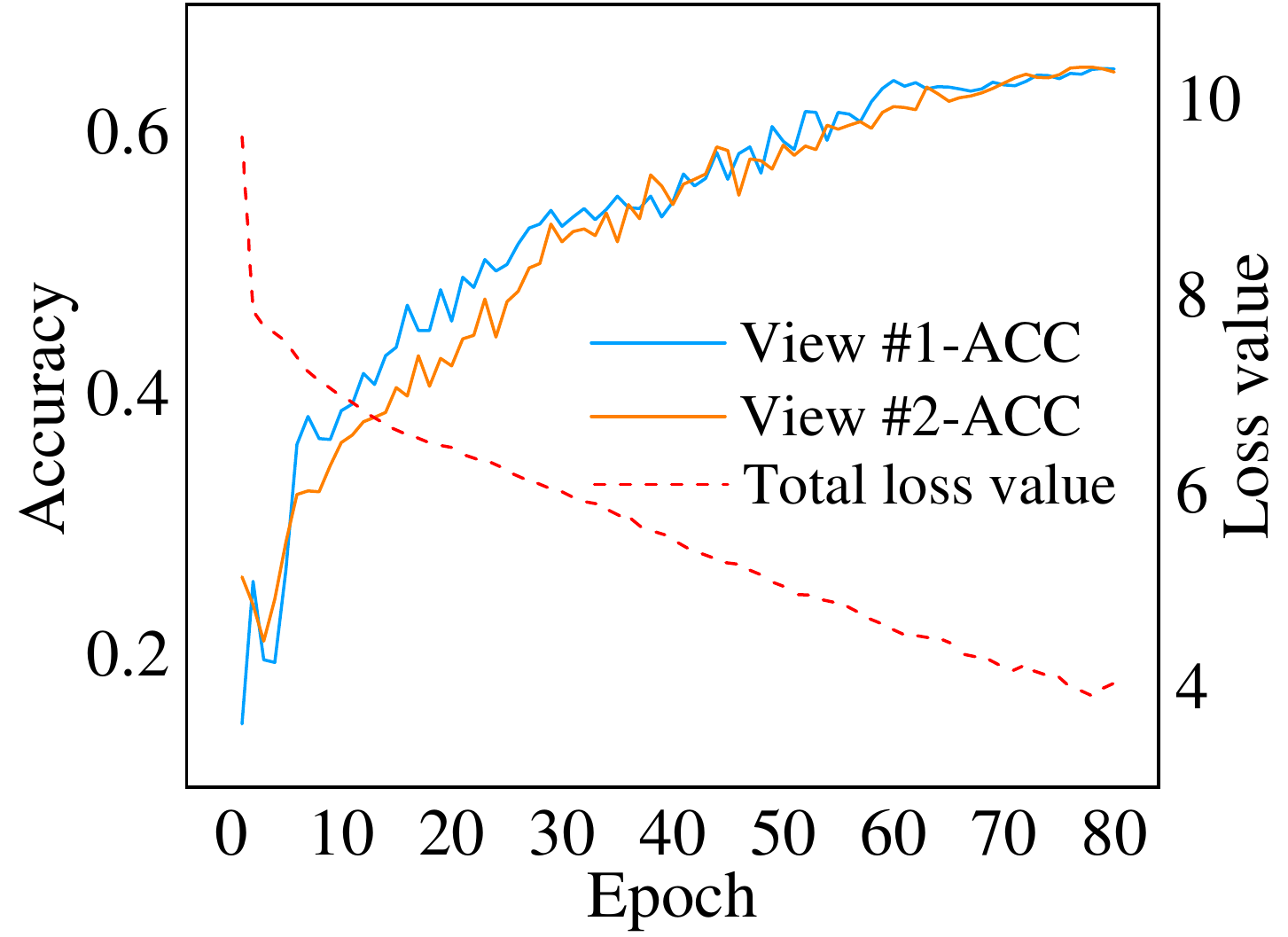}
}
\quad
\subfigure[Convergence on BDGP]{
\includegraphics[width=3.9cm]{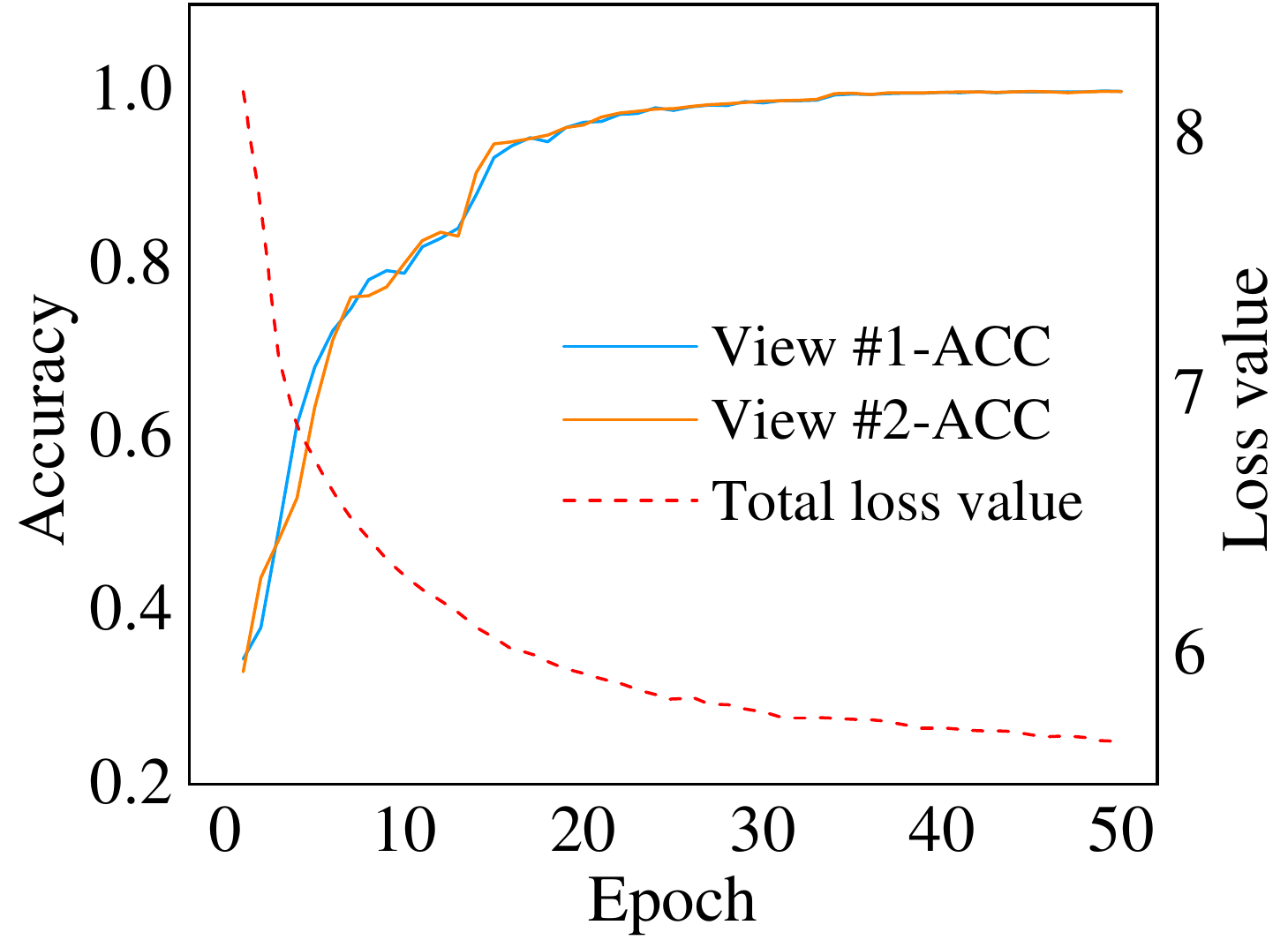}
}
\quad
\subfigure[Convergence on Caltech-5V]{
\includegraphics[width=3.9cm]{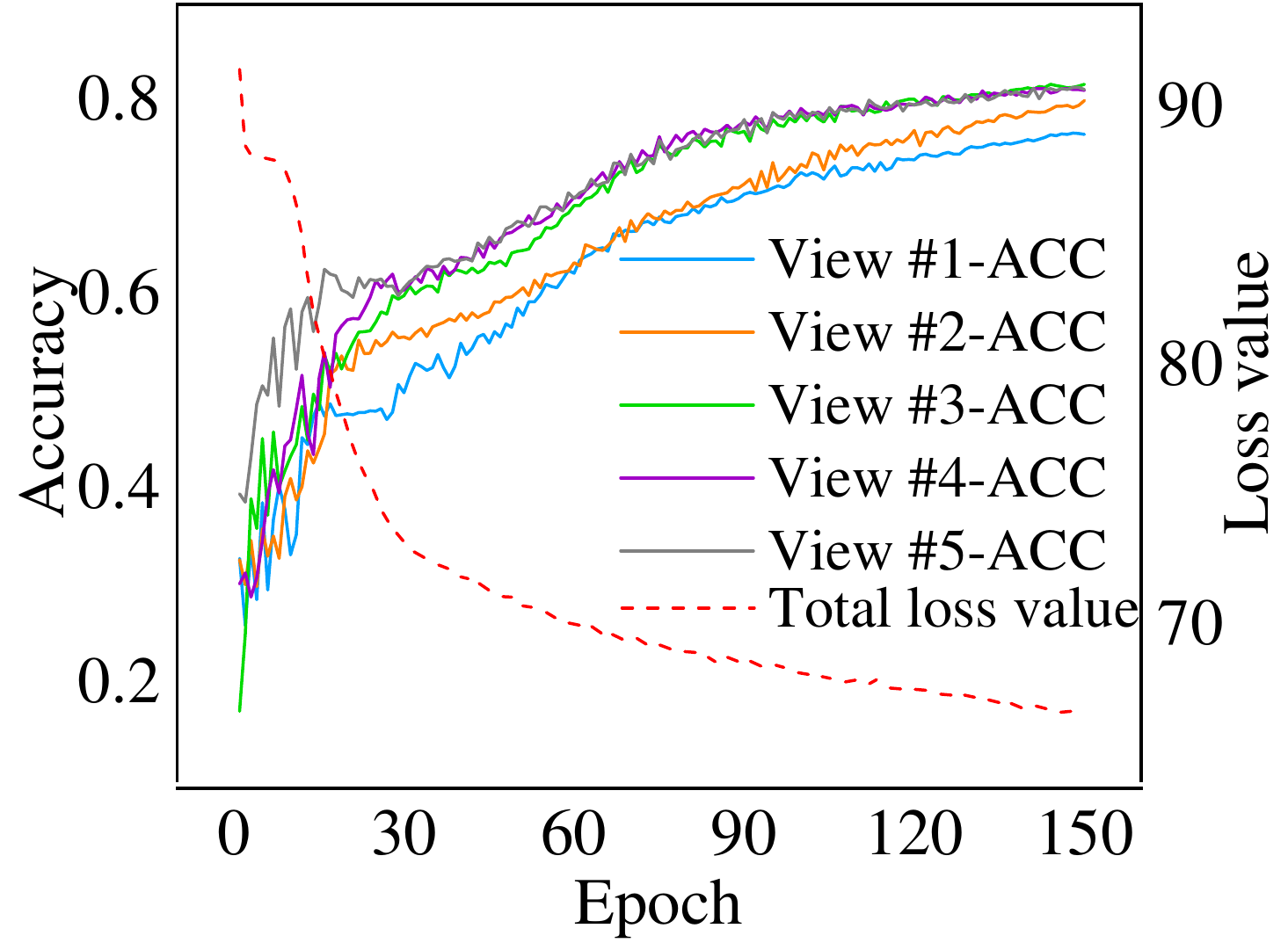}
}
\quad
\subfigure[Convergence on Fashion]{
\includegraphics[width=3.9cm]{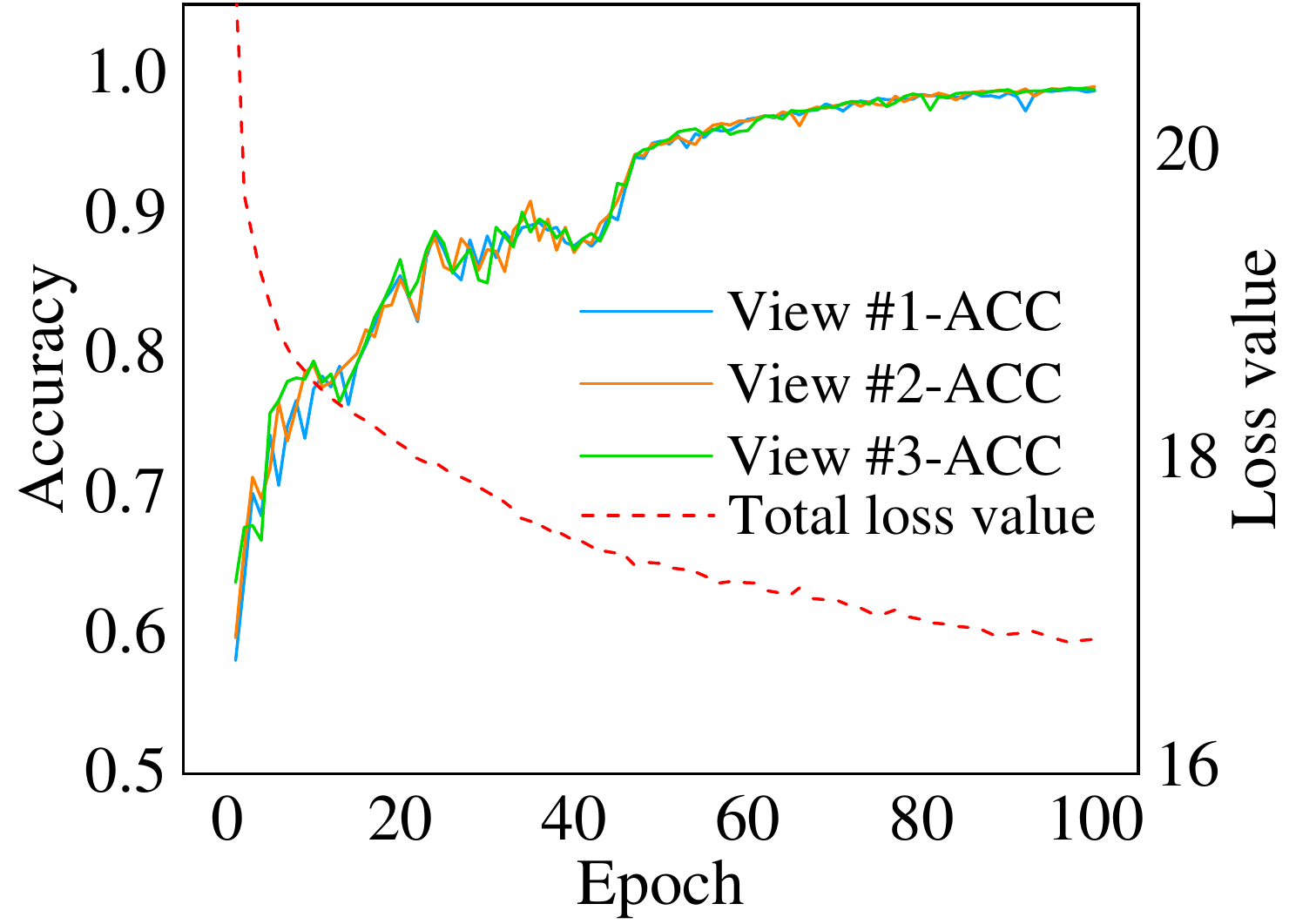}
}

\caption{Clustering accuracy of DistilMVC. The x-axis denotes the training epoches on four datasets, the left and right y-axis denote the clustering accuracy and corresponding loss value, respectively.}
\label{fig7}
\end{figure*}

\begin{figure*}[htbp]
\centering
\subfigure[3D bar graph]{
\includegraphics[width=4.8cm]{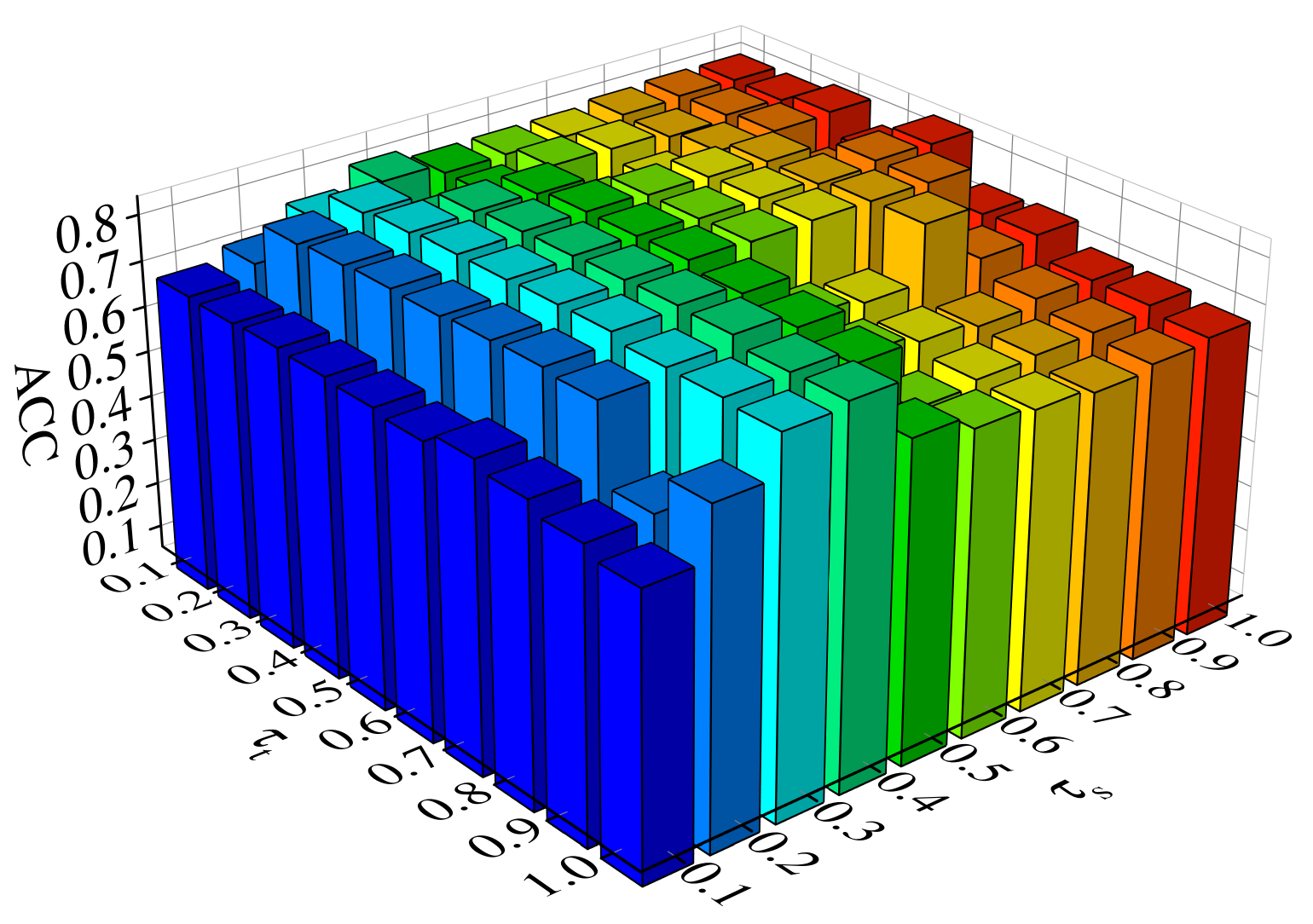}
}
\quad
\subfigure[3D surface graph]{
\includegraphics[width=4.8cm]{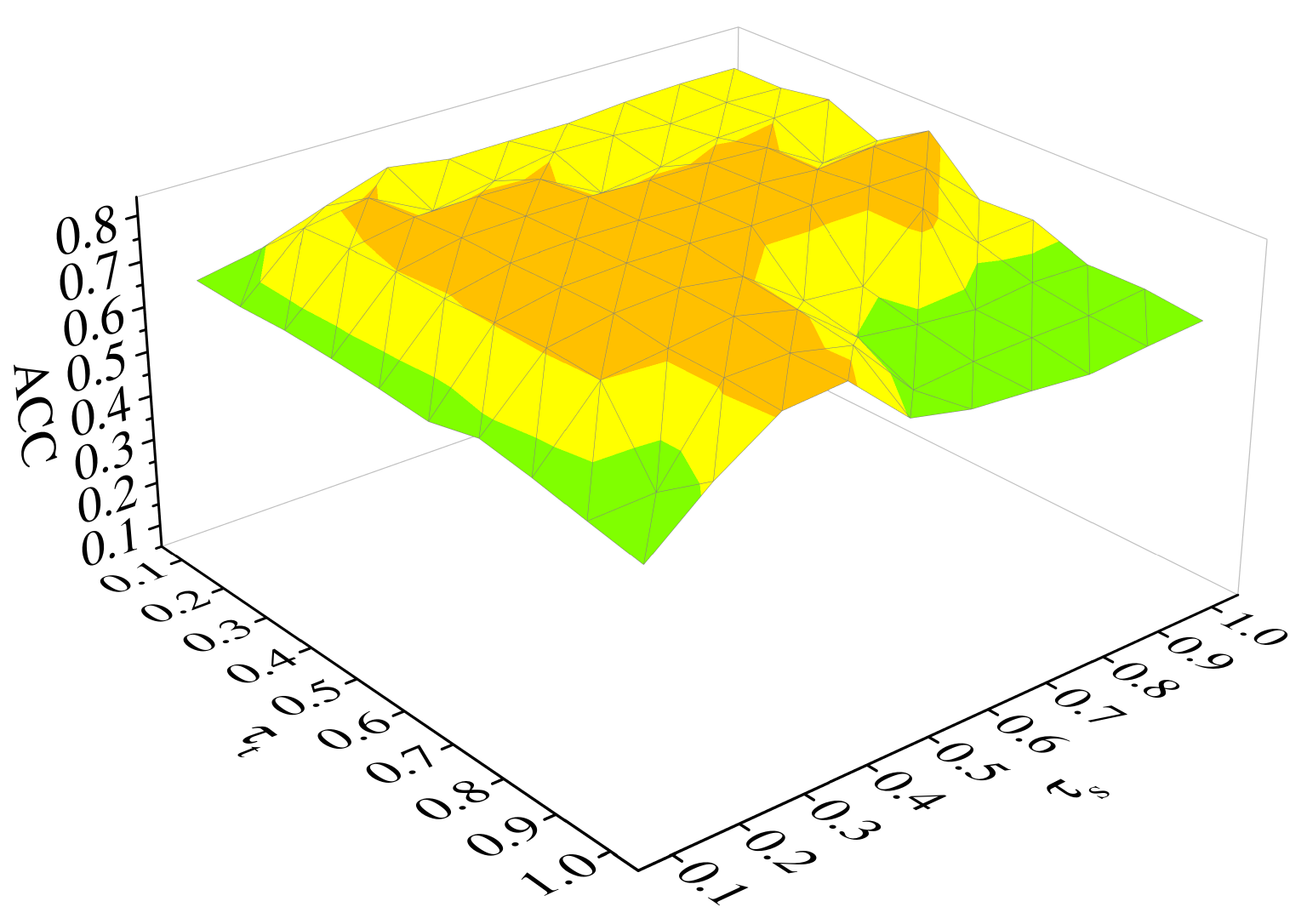}
}
\quad
\subfigure[Top view of 3D surface graph]{
\includegraphics[width=4.8cm]{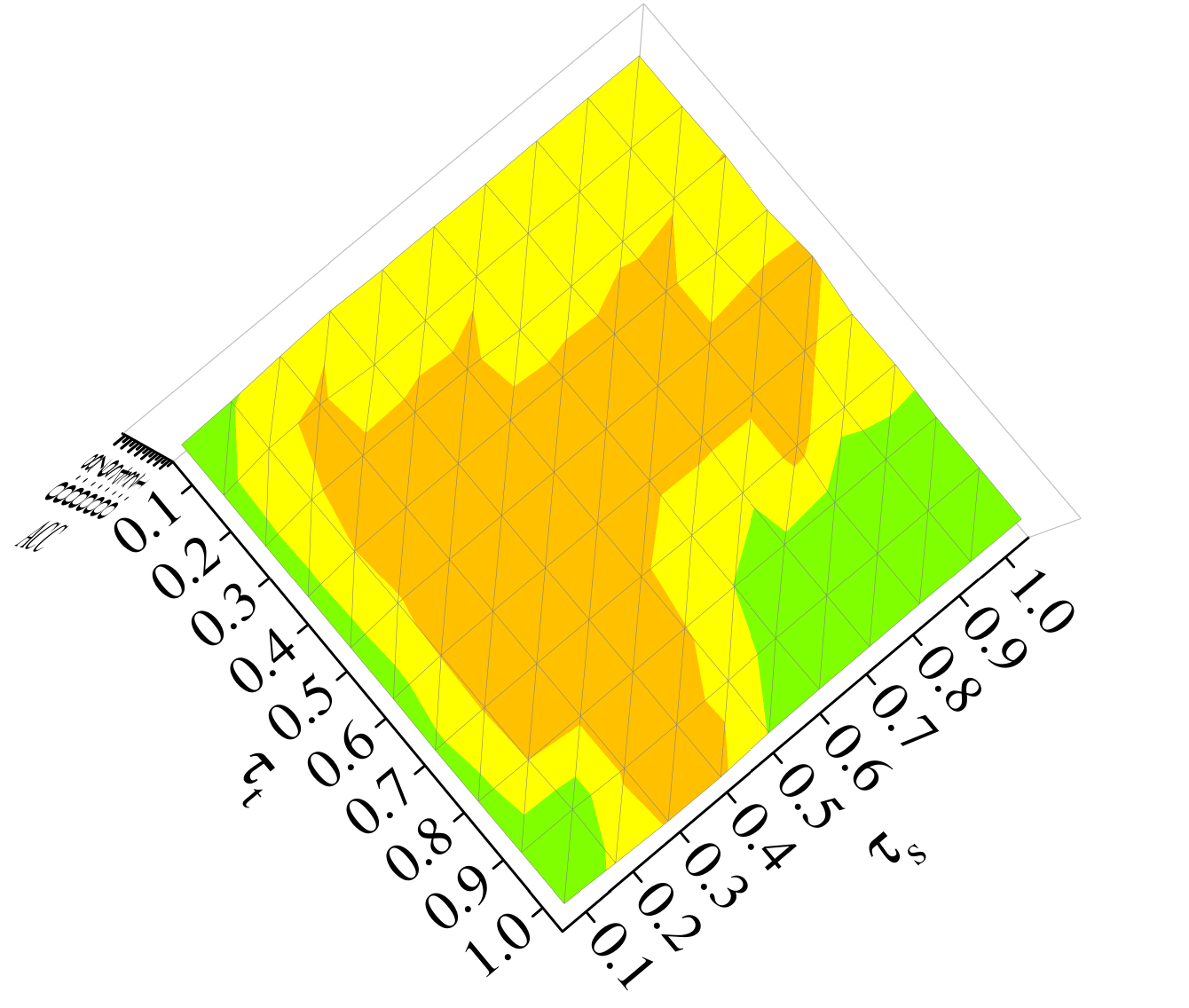}
}
\caption{The clustering performance of DistilMVC on the Caltech-5V dataset with different parameters $\tau_{s}$ and $\tau_{t}$, including 3D bar graph (a) and 3D surface graph (b,c). In the 3D surface graphs (b,c), the green region, yellow region, and orange region indicate that the ACC is in the ranges $(0.6, 0.7]$, $(0.7, 0.8]$, and $(0.8, 0.9]$, respectively.}
\label{fig8}
\end{figure*}

\begin{figure}[htbp]
\centering
\includegraphics[width=1\linewidth]{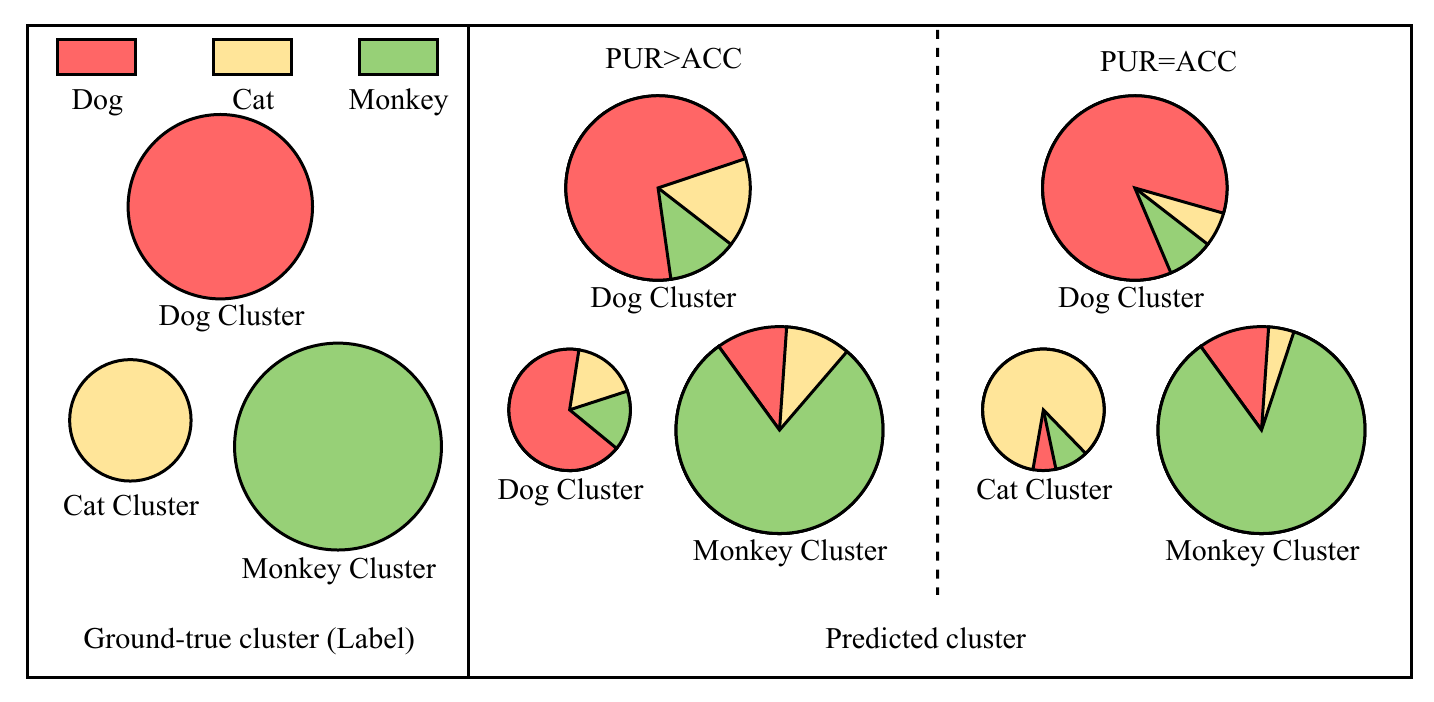}
\caption{The relation between PRU and ACC values. PUR$=$ACC indicates that there is a one-to-one correspondence between the predicted labels of clusters and their ground-true labels. When PUR$>$ACC, there exists duplicated clusters.
Since the proportion of "dog" in the predicted clusters is larger, there are two clusters marked with the label of "dog".}
\label{fig5}
\end{figure}

Since the over-confident pseudo-labels generated by baselines provide incorrect clustering directions.
On the other hand, DistilMVC use dark knowledge instead of pseudo-labels to provide more precise guide for self-supervised clustering, and thus correct the false clustering directions, while using the Hungarian algorithm to ensure that the label of each cluster is distinct. So the ground-ture cluster labels and predicted cluster labels have one-to-one correspondence. This is the core idea of multi-view self-distillation.

Unlike traditional and existing deep MVC approaches, our DistilMVC targets to further optimize the pseudo-labels learning. The overconfidence of pseudo-labels is alleviated by self-distillation, and robust clustering results are obtained by learning different hierarchies of mutual information to enforce the consistency of different views. In addition to the clustering performance, the visualization of the learned available features is shown in Fig. \ref{fig6}.
All datasets except Caltech-2V eventually converge well, and Caltech-2V has poor clustering due to its large number of views and small number of samples. We also find that the data distribution becomes more compact and independent through training, and the clustering density is higher, indicating that our multi-view self-distillation method achieves an effective improvement in clustering performance.

\subsection{Model Analysis}

\subsubsection{Convergence Analysis}

We investigate the convergence of DistilMVC by reporting the loss value and the corresponding clustering performance with increasing epochs. As shown in Fig. \ref{fig7}, one could observe that the loss remarkably decreases in the first 20 epochs, and meanwhile, the ACC of different views continuously increases and tends to be smooth and consistent.

\subsubsection{Parametric Analysis}

The temperature hyperparameters $\tau_{s}$ (Formula \ref{eq4}) and $\tau_{t}$ (Formula \ref{eq7}) are used to control the shape of the distribution. As shown in Fig. \ref{fig8} (a) we change their values in the range of $[0.1, 1.0]$ and the interval is 0.1. 

In Fig. \ref{fig8} (c), the orange region belongs to the temperature comfort zone, accounting for 37.04\% of the total region and is in the center. The dark knowledge in this region contains rich semantic information, i.e., the KL divergence between the dark knowledge and the output distribution of the student network is lower, which also proves that DistilMVC can bring high-quality supervision to the student network. The yellow and green regions account for 45.73\% and 17.23\% of the total region, respectively, and are distributed at the edges. The yellow region is between the orange region and the green region, which is a buffer zone, and the clustering performance decreases slightly in this region. The green region proves that the temperature $\tau_{s}$ and $\tau_{t}$ are too large or too small, which will obviously reduce the clustering performance, so our choice needs to avoid the green region. The reasons are as follows: (1) When $\tau_{s}$ and $\tau_{t}$ are close to 1 at the same time, they will enter the green region. The reason is that the temperature $\tau_{s}$ and $\tau_{t}$ are too large and the distribution is too smooth, so that the model fails to learn the focus and collapses. (2) When $\tau_{t}$ is 0.1, it will enter the green region. The reason is that the temperature $\tau_{t}$ is too small and the distribution is too peak, so the model will pay special attention to difficult negative samples, making it difficult for the model to converge or the learned features to generalize.

\subsubsection{Ablation experiment}

We perform the ablation study to demonstrate the importance of each component of our method. As shown in Table \ref{table4}, we designed six sets of schemes on four datasets with different numbers of views and observed the following results: a) All losses play an integral role in DistilMVC; b) A significant improvement is obtained after introducing the self-distillation method on (1)(3)(5)(6), which further proves that our method can effectively mitigate the problem of overconfidence in pseudo-labels and thus improve the clustering performance; c) The addition of self-distillation in (2)(4) leads to model degradation;  d) Comparing (1) and (6) we can see that optimizing the loss $\mathcal{L}_{con}$ can lead to a huge improvement, proving the effectiveness of our proposed method for maximizing mutual information at different hierarchies; e) The above four observations hold for all data sets, which also demonstrates the robustness of our method.

The reasons for the above observations can be explained as follows: a) $\mathcal{L}_{rec}$ establishes the feature space for feature learning, $\mathcal{L}_{con}$ learns features by maximizing mutual information at different hierarchies, and $\mathcal{L}_{self}$ improves error prediction by reducing the confidence of the model, and each of the three components is responsible for and reinforces each other. b) The pseudo-labels are derived from the high-dimensional features learned by the teacher network, and the self-distillation method can transform the pseudo-labels into dark knowledge, improving the quality of the supervised signal. c) View reconstruction is conducive to maintaining the complementarity between views, which is the basis of feature learning. If $\mathcal{L}_{rec}$ is skipped and $\mathcal{L}_{con}$ is directly optimized, complementary information will be lost. Therefore, for (2), the features learned by the teacher network are not linearly separable due to the lack of complementary information, so they are not suitable for distillation. For (4), teacher networks are not involved in learning, and blind distillation can provide more false labels to student networks. d) Optimized $\mathcal{L}_{con}$ is able to maximize mutual information at different hierarchies from teacher, student, and encoder, which greatly facilitates consistent learning. e) DistilMVC has strong generalization ability and robustness. Thus multi-view self distillation is well suited for feature learning and clustering in stages for high qualified clustering.

\begin{table*}[htbp]
\caption{Ablation studies on loss components on Caltech-2V,  Caltech-3V, Caltech-4V and Caltech-5V. "\checkmark" denotes DistilMVC with the component, and "*" indicates the method of adding self-distillation on the original model.}
\label{table4}
\renewcommand\arraystretch{1}
\centering
\setlength{\tabcolsep}{0.6mm}{
\begin{tabular}{lccccccccccccccccc}
\hline
& \multirow{2}{*}{$\mathcal{L}_{rec}$} & \multicolumn{3}{c}{$\mathcal{L}_{con}$} & \multirow{2}{*}{$\mathcal{L}_{self}$} & \multicolumn{3}{c}{Caltech-2V}  & \multicolumn{3}{c}{Caltech-3V} & \multicolumn{3}{c}{Caltech-4V} & \multicolumn{3}{c}{Caltech-5V} \\
&    & $\mathcal{L}_{tea}$ & $\mathcal{L}_{stu}$ & $\mathcal{L}_{IIC}$ &
& ACC    & NMI    & PUR    & ACC    & NMI    & PUR    & ACC    & NMI    & PUR    & ACC    & NMI    & PUR    \\ \hline
(1)  & \checkmark  &  &  &  &  & 0.3043   & 0.1965   & 0.3043   & 0.1614   & 0.0162   & 0.1614   & 0.1429   & 0.0043   & 0.1429   & 0.1429   & 0.0101   & 0.2450   \\
(1*) & \checkmark  &  &  &  & \checkmark  & 0.4886   & 0.3099   & 0.5036   & 0.4736   & 0.3285   & 0.4993   & 0.4621   & 0.3634   & 0.4786   & 0.4914   & 0.3798   & 0.5243   \\ \hline
(2)  &  & \checkmark   & \checkmark  & \checkmark  & & 0.5286   & 0.4365   & 0.5464   & 0.5071   & 0.4357   & 0.5114   & 0.5393   & 0.4395   & 0.5279   & 0.7350   & 0.5902   & 0.7350   \\
(2*) &  & \checkmark   & \checkmark  & \checkmark  & \checkmark & 0.5136   & 0.4513   & 0.5414   & 0.4793   & 0.4461   & 0.5129   & 0.5086   & 0.4954   & 0.5400   & 0.7321   & 0.5910   & 0.7921   \\ \hline 
(3)  & \checkmark  & \checkmark  &  & \checkmark  &  & 0.3864   & 0.3090   & 0.3864   & 0.1429   & 0.0009   & 0.1429   & 0.1436   & 0.0018   & 0.1436   & 0.3543   & 0.2414   & 0.3657   \\
(3*) & \checkmark  & \checkmark  &  & \checkmark  & \checkmark  & 0.5507   & 0.4472   & 0.5514   & 0.5871   & 0.5175   & 0.5921   & 0.6271   & 0.5768   & 0.6271   & 0.7600   & 0.6929   & 0.7600   \\ \hline
(4)  & \checkmark  &  & \checkmark  & \checkmark  &  & 0.5650   & 0.5033   & 0.5871   & 0.6200   & 0.5270   & 0.6286   & 0.7250   & 0.6528   & 0.7350   & 0.7643   & 0.6904   & 0.7643   \\
(4*) & \checkmark  &  & \checkmark  & \checkmark  & \checkmark  & 0.5621   & 0.5214   & 0.5686   & 0.5836   & 0.5039   & 0.6029   & 0.6671   & 0.6158   & 0.6821   & 0.7443   & 0.6522   & 0.7443   \\ \hline
(5)  & \checkmark  & \checkmark  & \checkmark  &  &  & 0.5814   & 0.5055   & 0.5921   & 0.6364   & 0.5654   & 0.6536   & 0.7971   & 0.6838   & 0.7971   & 0.7971   & 0.6838   & 0.7971   \\
(5*) & \checkmark  & \checkmark  & \checkmark  &  & \checkmark  & 0.5843   & 0.5327   & 0.5864   & 0.6371   & 0.5649   & 0.6543     & 0.8057   & 0.6954   & 0.8057   & 0.8057   & 0.6954   & 0.8057   \\ \hline
(6)  & \checkmark  & \checkmark  & \checkmark  & \checkmark  &  & 0.5779   & 0.4958   & 0.5921   & 0.6343   & 0.5659   & 0.6536   & 0.7993   & 0.6863   & 0.7993   & 0.8171   & 0.6930   & 0.8171   \\
(6*) & \checkmark  & \checkmark  & \checkmark  & \checkmark  & \checkmark  & 0.6192   & 0.5329   & 0.6192   & 0.6500   & 0.5751   & 0.6629   & 0.8086   & 0.6951   & 0.8086   & 0.8236   & 0.7090   & 0.8239   \\ \hline
\end{tabular}
}
\end{table*}

\section{Conclusion}
In this paper, we propose a novel and flexible DistilMVC, which can handle all kinds of multi-view data to enable effective multi-view clustering. Based on a self-distilled architecture, DistilMVC can effectively alleviate false predictions caused by the overconfidence in pseudo-labels, and when combined with a feature learning method of different hierarchies of mutual information, it achieves SOTAs on eight datasets. Thus, it solves a persistent nuisance of MMVC: the pseudo-labels obtained by feature learning are not adequate for self-supervised signals. Such a unified framework will provide novel insight for the community to understand multi-view clustering. 
In the future, we plan to further explore the potential of our theory and framework for other multi-view learning tasks, such as incomplete multi-view clustering, cross-modal retrieval, and 3D reconstruction.


\begin{IEEEbiography}
[{\includegraphics[width=1in,height=1.25in,clip,keepaspectratio]{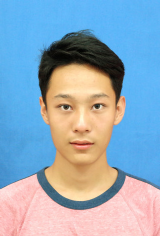}}]{Jiatai Wang}
received M.S. degree from Inner Mongolia University of Technology, Hohhot, China. Recently, he is working as a visiting scholar and going to pursue his Ph.D degree at Nankai University, Tianjin, China. He has published several papers in high impact journals in computer vision field, such as IET computer vision. His interests are focused on unsupervised learning in CV field. 
\end{IEEEbiography}

\begin{IEEEbiography}
[{\includegraphics[width=1in,height=1.25in,clip,keepaspectratio]{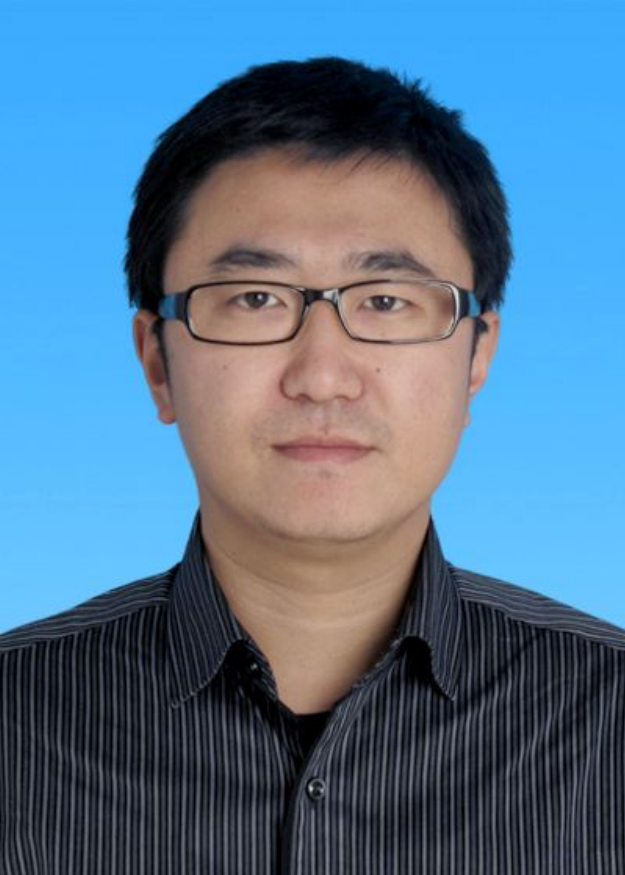}}]{Zhiwei Xu}
received the B.S. degree in 2002 from University of Electronic Science and Technology of China, Chengdu, China, and the Ph.D. degree in 2018 from Institute of Computing Technology, Chinese Academy of Sciences, Beijing, China. He is an associate professor and M.S. supervisor of Inner Mongolia University of Technology, while working as an Adjunct Professor in Institute of computing, Chinese Academy of Sciences. From 2020 to 2021, he worked towards visiting post-doctoral in the Department of Electrical and Computer Engineering, State University of New York at Stony Brook, Stony Brook, NY. His research interests include in-network data compact representation, learning, and the related security and privacy problems.
\end{IEEEbiography}

\begin{IEEEbiography}
[{\includegraphics[width=1in,height=1.25in,clip,keepaspectratio]{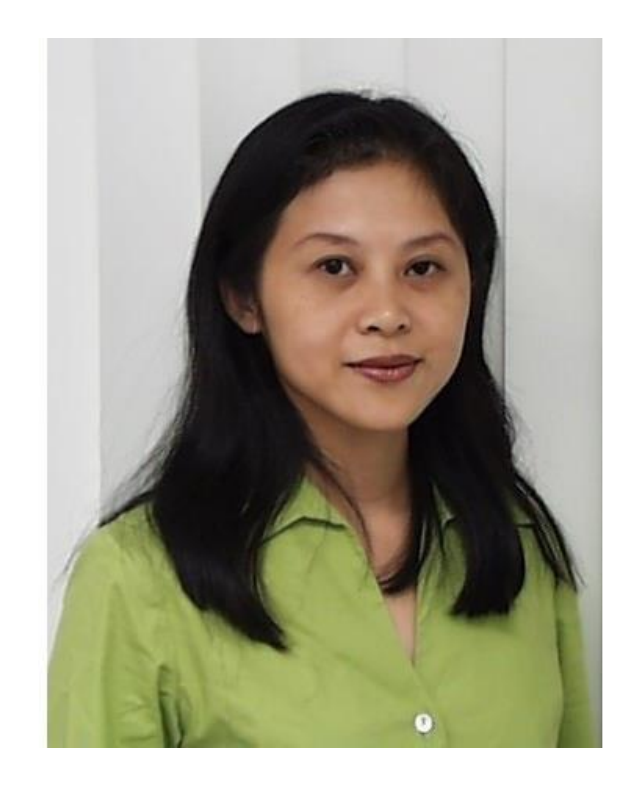}}]{Xin Wang}
received the B.S. and M.S. degrees in telecommunications
engineering and wireless communications engineering respectively from Beijing University of Posts and Telecommunications, Beijing, China, and the Ph.D. degree in electrical and computer engineering from Columbia University, New York, NY. She is currently an Associate Professor in the Department of Electrical and Computer Engineering of the State University of New York at Stony Brook, Stony Brook, NY. Before joining Stony Brook, she was a Member of Technical Staff in the area of mobile and wireless networking at Bell Labs Research, Lucent Technologies, New Jersey, and an Assistant Professor in the Department of Computer Science and Engineering of the State University of New York at Buffalo, Buffalo, NY. Her research interests include algorithm and protocol design in wireless networks and communications, mobile and distributed computing, as well as big data analysis and machine learning. She has served in executive committee and technical committee of numerous conferences and funding review panels, and serves as the associate editor of IEEE Transactions on Mobile Computing. Dr. Wang achieved the NSF career award in 2005, and ONR challenge award in 2010.
\end{IEEEbiography}

\end{document}